\documentclass[reprint,aps,pre]{revtex4-2}

\usepackage{amsmath,amsfonts,amscd,amssymb,graphicx,subeqnar}

\usepackage{dcolumn}
\usepackage{bm}
\usepackage{upgreek}
\usepackage[percent]{overpic}
\usepackage{color}
\usepackage{adjustbox}
\usepackage{rotating}
\usepackage{float}
\makeatletter
\let\newfloat\newfloat@ltx
\makeatother
\usepackage{algorithm}
\usepackage{algpseudocode}
\algnewcommand\algorithmicinput{\textbf{Input:}}
\algnewcommand\Input{\item[\algorithmicinput]}
\usepackage{soul}
\usepackage{stackengine}

\usepackage{tikz-cd} 

\usepackage[hypertexnames=false,
            colorlinks=true,
            linkcolor=blue,
            citecolor=blue,
            urlcolor=blue]
            {hyperref} 

\usepackage{tikz}
\definecolor{red}{rgb}{0.8500, 0.3250, 0.0980}
\definecolor{green}{rgb}{0.4660, 0.6740, 0.1880}
\definecolor{yellow}{rgb}{0.9290, 0.6940, 0.1250}
\definecolor{blue}{rgb}{0, 0.4470, 0.7410}
\usepackage{titlesec}
\usepackage{colortbl}
\titleformat*{\subsection}{\bfseries\raggedright}
\usepackage{makecell}
\usepackage{hyperref}
\begin{document}

\title{Inverse Design with Dynamic Mode Decomposition}

\author{Yunpeng Zhu$^{1,*}$, Liangliang Cheng$^{2,*}$, Anping Jing$^{3}$, Hanyu Huo$^{4}$, Ziqiang Lang$^{5}$, Bo Zhang$^{3}$, J. Nathan Kutz$^{6,*}$}

\affiliation{$^{1,*}$School of Engineering and Material Science, Queen Mary University of London, London, E1 4NS, UK}
\affiliation{$^{2}$ENTEG, Faculty of Science and Engineering, University of Groningen, Groningen, 9747 AG, The Netherlands}
\affiliation{$^{3}$School of Mechanical Engineering, Ningxia University, Yinchuan, 750021, China}
\affiliation{$^{4}$School of Chemistry and Material Science, University of Science and Technology of China, Hefei, 230026, China}
\affiliation{$^{5}$\mbox{School of Electrical and Electronic Engineering, The University of Sheffield, Sheffield, S10 2TN, UK}}
\affiliation{$^{6,*}$Department of Applied Mathematics and Electrical and Computer Engineering, University of Washington, Seattle, WA USA} %
\email{yunpeng.zhu@qmul.ac.uk; \\ liangliang.cheng@rug.nl; \\ kutz@uw.edu} 

\begin{abstract}
We introduce a computationally efficient method for the automation of inverse design in science and engineering.  Based on simple least-square regression, the underlying dynamic mode decomposition algorithm can be used to construct a low-rank subspace spanning multiple experiments in parameter space.  The proposed {\em inverse design dynamic mode composition} (ID-DMD) algorithm leverages the computed low-dimensional subspace to enable fast digital design and optimization on laptop-level computing, including the potential to prescribe the dynamics themselves.  Moreover, the method is robust to noise,  physically interpretable, and can provide uncertainty quantification metrics.  The architecture can also efficiently scale to large-scale design problems using randomized algorithms in the ID-DMD. The simplicity of the method and its implementation are highly attractive in practice, and the ID-DMD has been demonstrated to be an order of magnitude more accurate than competing methods while simultaneously being 3-5 orders faster on challenging engineering design problems ranging from structural vibrations to fluid dynamics. Due to its speed, robustness, interpretability, and ease-of-use, ID-DMD in comparison with other leading machine learning methods represents a significant advancement in data-driven methods for inverse design and optimization, promising a paradigm shift in how to approach inverse design in practice.
\end{abstract}

\maketitle


\section{INTRODUCTION} \label{intro}

Two of the most common tasks in engineering revolve around control and design. For inverse design, which is the focus of our work, empirical methods have been used since the invention of tools, i.e. trial-and-error and experimentation. Design of experiments~\cite{fisher1966design,martins2021engineering} was the first formal mathematical architecture and principled approach to the inverse design problem. Early efforts in design used governing equations, design principles and experiments in an integrated fashion to enact the design of experiments. In the last two decades, this process has been accelerated through advanced computational techniques (i.e. finite element methods) and their reduced order modeling counterparts~\cite{cook2007concepts,touze2021model}. More recently, data-driven methods and machine learning have allowed for training models on high-fidelity simulations in order to accelerate inverse design~\cite{Billings2003design,lu2021physics,li2023fourier,allen2022physical}. In this case, the training time can be computationally intensive, but deployment can then be efficient. Here, we introduce a new data-driven strategy which is a direct least-square regression to an optimized system model, i.e. there is no expensive training. The proposed {\em inverse design dynamic mode composition} (ID-DMD) algorithm leverages a computed low-dimensional linear subspace to enable fast digital design and optimization which is orders of magnitude faster than competing data-driven and computational design methods. ID-DMD is a significant new inverse design paradigm which shapes the design process and workflow 
due to its demonstrated robustness, computational efficiency, scalability, ease of implementation, and interpretability.

Inverse design involves optimizing design parameters or system characteristics to meet specific performance goals. Over a long period of time, inverse design predominantly relied on empirical methods supported by the design of experiments~\cite{martins2021engineering}, established design principles~\cite{yan2019aerodynamic}, and governing equations, such as {\em Ordinary Differential Equations} (ODEs) and {\em Partial Differential Equations} (PDEs)~\cite{leveque2007finite}. However, these early techniques are either expensive or of low fidelity. Advanced computational techniques (i.e. finite difference, finite element, and boundary element)~\cite{gupta2022finite} have since addressed many of these shortcomings by enabling high-fidelity numerical simulations of complex systems. Accurate simulations allow designers to leverage optimal search algorithms~\cite{deb2012optimization}, such as Newton’s methods, Lagrange Multipliers, and evolutionary algorithms, to achieve target system properties within a given design space. Although reduced order models accelerate numerical simulations through dimensional reduction approaches, such as Principal Component Analysis (PCA)~\cite{lang2009reduced} and Proper Orthogonal Decomposition (POD)~\cite{ghoman2012pod}, inverse design based on numerical simulations remains inherently resource-intensive and time-consuming, especially for large-scale, multi-physics, and highly complex systems.

  \begin{figure*}[t]
      \centering
      \includegraphics[width=\textwidth]{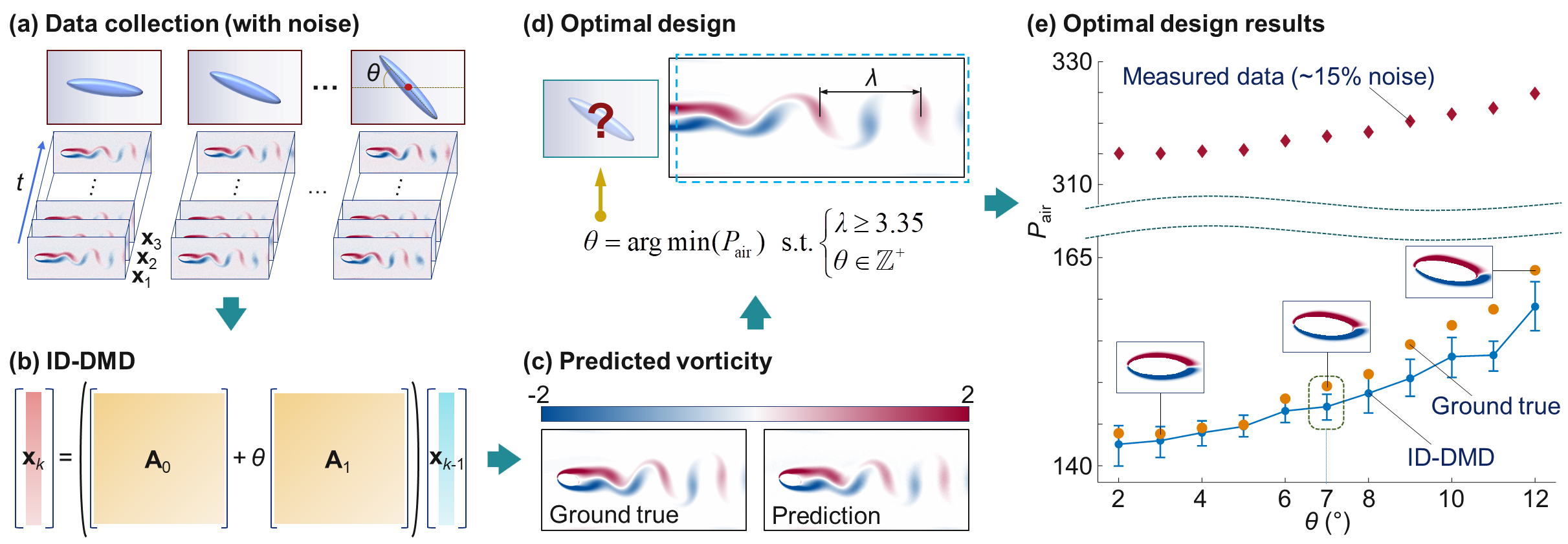}
      \caption{
     The inverse design of airfoil pitched angle using ID-DMD. (a) Collect vorticity simulation data as snapshots across varying pitched angles $\theta$, with additional noise levels up to 15\%. (b) Identify ID-DMD to capture the underlying dynamics of the system. (c) Perform model validation by evaluating vorticity predictions at different pitched angles. (d) Utilize the validated ID-DMD for airfoil pitched angle optimization. The design target is to achieve a preferred wavelength $\lambda$, while minimizing the airflow power ${P}_\text{air}$ across the fluid filed within the blue dotted box. (e) The design is robust to noise and stable with narrow uncertainty boundaries.
      }
      \label{fig_1}
  \end{figure*}

The landscape of inverse design recently experienced a significant shift with the emergence of data-driven approaches and scientific machine learning. 
%
Advanced machine learning and AI algorithms are empowering machine intelligence broadly across science and engineering disciplines. Such algorithms are leveraging both data and computations in order to train algorithms which are currently being deployed and tested in real-world environments. The majority of these technologies have revolved around discovery, forecasting (time-series) and control. Typical machine learning approaches, such as Convolutional Neural Networks (CNN)~\cite{gu2018recent}, deep reinforcement learning~\cite{sutton2018reinforcement} and deep Model Predictive Control (MPC)~\cite{lenz2015deepmpc}, improve decision-making and system control in games~\cite{silver2018general}, robotics~\cite{kober2013reinforcement}, industrial automation~\cite{salzmann2023real}, and process control~\cite{lenz2015deepmpc}. Moreover, generative learning approaches that include Autocoder-based methods (latent space roll outs)~\cite{schmidt2009distilling} and
transformer and foundation models~\cite{khan2022transformers,bommasani2021opportunities} enable large data analysis for language processing~\cite{wolf2020transformers}, computer vision~\cite{khan2022transformers}, and speech recognition~\cite{kim2022squeezeformer}. However, all these algorithms require large training data sets and are rarely applied to system designs. Further, they are expensive to train and largely black-box.

In regard to design, machine learning and AI algorithms develop surrogate models that act as efficient simulators, replacing complex numerical simulations to expedite the inverse design process. A dominant approach within this domain is operator learning, which includes approaches such as Physics-Informed Neural Networks (PINNs)~\cite{lu2021physics}, Deep Operator Networks (DeepONet)~\cite{lu2022multifidelity}, Neural Implicit Flow (NIF)~\cite{pan2023neural} and Fourier Neural Operators (FNOs)~\cite{li2023fourier}. One may employ PINNs as a solver when the governing differential equations of a system are known in advance. In contrast, DeepONet, NIF, and FNO were designed to learn system dynamic responses, purely rely on data, and function as a simulator. Specifically, DeepONet introduces a unique architecture comprising two sub-networks: the branch net, which processes input functions and initial conditions, and the trunk net, which handles output locations and design parameters~\cite{shukla2024deep}. This framework allows DeepONet to predict system dynamics by integrating information from the branch and trunk nets, enabling applications in system design. The NIF extends DeepONet by utilizing fully connected architectures for both the branch and trunk networks, enhancing its flexibility and generalization capabilities. Furthermore, FNO incorporates the Fourier transform into its network architecture, enabling it to learn resolution-invariant operators~\cite{kovachki2021universal}. In its continuous form, FNO can be seen as a special case of DeepONet~\cite{lu2022comprehensive}. Recently, DeepMind introduced a data-driven inverse design method using Graph Neural Networks (GNN)~\cite{allen2022physical}. The GNN-based design integrates finite element meshes with GNN to facilitate efficient geometry and shape optimization for fluid dynamic systems. The existing data-driven inverse design approaches have demonstrated efficiency in inverse design once the networks are properly trained~\cite{zhang2024blending}. However, their complex structures often lead to slow training and simulation processes using high-performance computers relying on GPU or TPU computations. Furthermore, the black-box nature of neural networks limits their physical interpretability, which can impede their ability to generate robust and reliable designs when working with noisy data.

Unlike conventional neural network-based data-driven methods, {\em Dynamic Mode Decomposition} (DMD)~\cite{brunton2022data} offers a distinctive network-free approach for capturing complex system dynamics through a linear operator approximation. The DMD enables a state-space representation of dynamic systems that can be readily applied for system analysis and control~\cite{tu2013dynamic,proctor2016dynamic,lusch2018deep,han2020dynamic}. Yet, no results have demonstrated the application of DMD in the inverse design of dynamic systems. In this work, we propose an ID-DMD algorithm that designers should first attempt on a personal device (i.e. laptop) before developing more complicated approaches.  
The algorithm, which is illustrated in Fig.\ref{fig_1} and detailed in the results, demonstrates the ID-DMD optimization for the shape of a simple pitched airfoil. 
The ID-DMD applies simple least-square regression to best-fit linear dynamics across different design parameters. It utilizes a low-rank subspace encompassing multiple experimental parametrizations to enhance computational efficiency. Moreover, the ID-DMD explicitly reconstructs system dynamics with stable modes to ensure reliability. The efficiency and accuracy of the ID-DMD allow the use of optimization algorithms for the fit-for-purpose inverse design in a similar way to numerical design. The results demonstrate that ID-DMD offers several advantages:

\noindent (i) The ID-DMD enables efficient CPU-based training that is orders of magnitude faster and an order more accurate than existing data-driven approaches. 

\noindent (ii) The ID-DMD is scalable in order to handle complex design challenges across different types of challenging data from structural vibrations to fluid dynamics.

\noindent (iii) The ID-DMD is physically interpretable, securing long-term predictions over time and extrapolation of design parameters outside the training range.

\noindent (iv) The ID-DMD demonstrates robustness to noise and can provide uncertainty quantification metrics using integrated bagging methods.

\noindent (v) The ID-DMD uniquely enables the design of system by prescribing intrinsic  dynamics rather than just output responses. 

\noindent
These demonstrated advantages in speed, robustness, interpretability, and ease-of-use, make ID-DMD a significant advancement in data-driven methods for inverse design and optimization, especially in comparison with other leading paradigms of PINNs, FNO, DeepONet, and NIF.

\section{RESULTS}

Denote  $\{\mathbf{x}_{k},\mathbf{x}_{k-1}\}\in \mathbb{R}^{m}$ is a pair of time-evolving snapshots that represent the system's states,  where $m$ is the dimension of the state vector,  $k$ represents the discrete time, $\mathbf{x}_{k}=\mathbf{x}(k\Delta t)$ with the sampling time $\Delta t$. The best-fit linear operator between the snapshot pairs, known as a DMD approximation, represents the system’s underlying dynamics by a linear model. In ID-DMD, the linear operator is parametrized as a linear-in-parameter matrix, such that the ID-DMD representation of a dynamic system is written as

\begin{equation}  
\begin{aligned}
  & \mathbf{x}_{k}=(\mathbf{A}_{0}+{\varepsilon_{1}}{\mathbf{A}_{1}}+{\varepsilon_{2}}{\mathbf{A}_{2}}+\cdots ){\mathbf{x}_{k-1}} \\ 
\end{aligned} \label{eq1}
\end{equation}

\noindent where $\mathbf{A}_{0},\mathbf{A}_{1},\mathbf{A}_{2},...\in {{\mathbb{R}}^{m\times m}}$ are linear operators that relate the two state vectors $\mathbf{x}_{k}$ and $\mathbf{x}_{k-1}$;  and $\varepsilon_{1},\varepsilon_{2},\ldots$ are the system design parameters. Eq.(\ref{eq1}) is also known as a parametric state space model~\cite{benner2015survey} in control theory for discrete-time systems.

The primary goal of the ID-DMD is to identify linear operators that comprehensively capture system dynamics, enabling accurate predictions of system responses across different design parameters. Here, we utilize a Singular Value Decomposition (SVD)-based dimensional reduction approach\cite{brunton2016compressed} to directly identify a low-rank sub-space representation of the high-dimensional ($m\times m$) linear operator.  We evaluate the dominant eigenvectors ${{\bm{\upphi}}_{j}}$ and eigenvalues ${{s}_{j}}$ of the linear operator ${\mathbf{A}_{0}}+{{\varepsilon}_{1}}{\mathbf{A}_{1}}+{{\varepsilon }_{2}}{\mathbf{A}_{2}}+\cdots$ , where $j\in {{\mathbb{Z}}^{+}}$ represents the order of the modes ranging from 1 to a sufficiently high rank (See Methods and \textcolor{blue}{Section 1 in Supplementary Materials}) so as to enable the prediction of the system states through

\begin{equation}  
\mathbf{x}_{k}=\sum\limits_{j\in {{\mathbb{Z}}^{+}}}{{{\bm{\upphi}}_{j}}{{\text{e}}^{{{s}_{j}}(k-1)}}{{b}_{j}}}=\mathbf{\Phi}\exp [\mathbf{S}(k-1)]\mathbf{b} \label{eq2}
\end{equation} 

\noindent where $\mathbf{b}={{\mathbf{\Phi}}^{\dagger}}{{\mathbf{x}}_{1}}$ with ${\mathbf{x}}_{1}$ representing the initial states of the system; $\mathbf{S}$ is a diagonal matrix containing the complex frequencies ${s}_{j}=\sigma_{j}+\text{j}{\omega_{j}}$ with $\omega_{j}$ being the frequency and $\sigma_{j}$ being the decay rate.

For nonlinear systems, we apply the Koopman operator theory\cite{nathan2018applied} to project the system’s state measurements into a higher-dimensional space. This projection enables the approximation of the system’s nonlinear dynamics in a linear space. In this case, the ID-DMD representation for nonlinear systems is:

\begin{equation}
\psi({{\mathbf{x}}_{k}})=({{\mathbf{A}}_{{\upkappa},{0}}}+{{\varepsilon }_{1}}{\mathbf{A}_{{\upkappa},{1}}}+{{\varepsilon }_{2}}{\mathbf{A}_{{\upkappa},{2}}}+\cdots)\psi({{\mathbf{x}}_{k-1}}) \label{eq3}
\end{equation}

\noindent where $\psi$ is known as observables mapping the states ${{\mathbf{x}}_{k}}$ to a high-dimensional space. The Koopman operators ${{\mathbf{A}}_{{\upkappa},{0}}},{\mathbf{A}_{{\upkappa},{1}}},{\mathbf{A}_{{\upkappa},{2}}},...$ are linear operators acting on the observables.

An optimization problem can then be formulated for the inverse design of either state-related performances (e.g., output amplitude, power) or intrinsic dynamic properties (e.g., natural frequencies, stability)~\cite{zhu2017design} as:

\begin{equation}
\bm{\upepsilon}_\text{d}=\arg \underset{{\bm{\upepsilon}_{\min}}\le \bm{\upepsilon}\le {{\bm{\upepsilon}}_{\max}}}{\mathop{\min}}\,L(\bm{\upepsilon})\ \ \ \ \text{s}\text{.t}\text{.}\ g(\bm{\upepsilon})\le 0,\ h(\bm{\upepsilon})=0  \label{eq4}
\end{equation}

\noindent where $\bm{\upepsilon}=\{{{\varepsilon}_{1}},{{\varepsilon}_{2}},\ldots \}$ is the design variable vector, $\bm{\upepsilon}_{\min}$ and $\bm{\upepsilon}_{\max}$ are the lower bounds and upper bounds of the design parameters, respectively. $\bm{\upepsilon}_\text{d}$ is the optimal design result. $L(\bm{\upepsilon})$ is the loss function for the optimal design problem.  $g(\bm{\upepsilon})$ and $h(\bm{\upepsilon})$ are inequality and equality constraints, respectively. Optimal searching algorithms can be applied to achieve optimal solutions tailored to the desired system performance. 

In the following studies, we will demonstrate that the ID-DMD can effectively represent a wide range of complex dynamic systems, including those with both low- and high-dimensional properties. The ID-DMD offers a significant advancement by simplifying the data-driven modeling process while maintaining high efficiency, interpretability and precision, making it a powerful tool for modern inverse design and optimization.

\subsection{Inverse Design of Airfoil Pitched Angle}

The airfoil geometry and shape optimization is a typical example of the inverse design of a complex dynamic system\cite{secanell2006design,de2019airfoil}, making it an ideal case to illustrate the overall design process using ID-DMD.  Here, we demonstrate the design of airfoil pitched angle $\theta$ through the ID-DMD in Fig.\ref{fig_1}.


In this example, the shape of the airfoil is simplified to an ellipse as demonstrated in \textcolor{blue}{Fig.S1}. The airflow around the ellipse is simulated at several different pitch angles across the design space ranging from $\theta \in [2, 12]^\circ$ . The training datasets for ID-DMD modeling are snapshots of the steady-state vorticity, spanning  $t\in [20, 50]\text{s}$ at uniformly distributed angles $\theta =\{2, 4, 6, 8, 10, 12\}^\circ$.  

We assume the target airflow dynamics exhibit minimum power ${P}_\text{air}$ with a specified wavelength requirement of $\lambda>3.35$. Here, ${P}_\text{air}$ is defined as the root mean value of the sum square vorticity over time. The optimal airfoil pitched angle that meets the required dynamic properties is obtained as $\theta=7^\circ$ by solving the optimization problem in  Fig.\ref{fig_1}(d). The predicted vorticities at the designed angle are shown in \textcolor{blue}{Fig.S2}. In addition, we also quantify the uncertainties of the ID-DMD-based design using the bagging approach\cite{sashidhar2022bagging}. The results show that (i) ID-DMD can reliably predict vorticities at any pitch angle within the design space, even in the presence of significant noise, and (ii) ID-DMD maintains stability with narrow uncertainty boundaries across the design space.

The ID-DMD approach significantly accelerates the inverse design process for complex dynamic systems. In this example, the greedy search design based on the ID-DMD ($\sim$30s) is about 100 times faster compared to that of finite element simulations executed via CPU computation (Intel Core i7 @ 1.20GHz) using MATLAB. Details of the modeling and design process are discussed in \textcolor{blue}{Section 2 in Supplementary Materials}.

\subsection{Applications to Challenging Dynamic Systems}
Tab.\ref{tab.1} applies the proposed ID-DMD to a wide range of challenging dynamic systems from physics to engineering applications. The selected examples are from simulation and experimental studies, covering linear and nonlinear dynamics and transient to periodic behaviors.

The selected systems in the first two rows of Tab.\ref{tab.1}, involving a nonlinearly damped building and a Van de Pol equation, are of low-dimension with the number of snapshots exceeding the number of states. The following examples from the incident jet to droplet control represent high-dimensional dynamic systems, where the number of snapshots is smaller than the number of states. Settings for the ID-DMD modeling are listed in \textcolor{blue}{Tabs.S1-S7}. The prediction results in Tab.\ref{tab.1}, as well as \textcolor{blue}{Figs.S3-S6}, show that the ID-DMD accurately reconstructs a wide range of complex dynamic systems under design parameters that were not seen in the training data.  Notably, the droplet control in the final row is an experimental study,  The experimental setup is shown in \textcolor{blue}{Fig.S7}, and the results illustrate the method’s potential for practical engineering applications. 

The approach can scale with randomized SVD algorithms~\cite{halko2011finding} or parallel QR~\cite{eiximeno2024pylom} to effectively study significantly high-dimensional systems, including 3D spatio-temporal dynamics and multi-modality data. Further details on the ID-DMD representation for all examples are provided in the \textcolor{blue}{Section 3 in Supplementary Materials}.

  \definecolor{Myellow}{rgb}{0.8,0.75,0.4}
  \linespread{1.5}
  \begin{table*}[!htb] 
    \begin{minipage}{\textwidth}
      \centering    
\noindent
\caption{Application of ID-DMD to complex dynamic systems}
\label{tab.1}

    \begin{tabular}{p{2.4cm} p{6.4cm} p{8.6cm}}

\rowcolor{Myellow!50}
\hline
\makecell[c] {Systems} & \makecell[c] {ID-DMD} & \makecell[c]{Validation} \\

\hline
\makecell[c]{Nonlinear \\ building}
& \makecell[l]{
$\psi (\mathbf{x}_{k})=(\mathbf{A}_{\upkappa,{0}}+{c_\text{non}}{\mathbf{A}_{\upkappa,{1}}})\psi (\mathbf{x}_{k-1})$ \\
$c_\text{non}$ - Nonlinear damping ratio
} & \makecell[c]{
\includegraphics[width=0.45\textwidth]{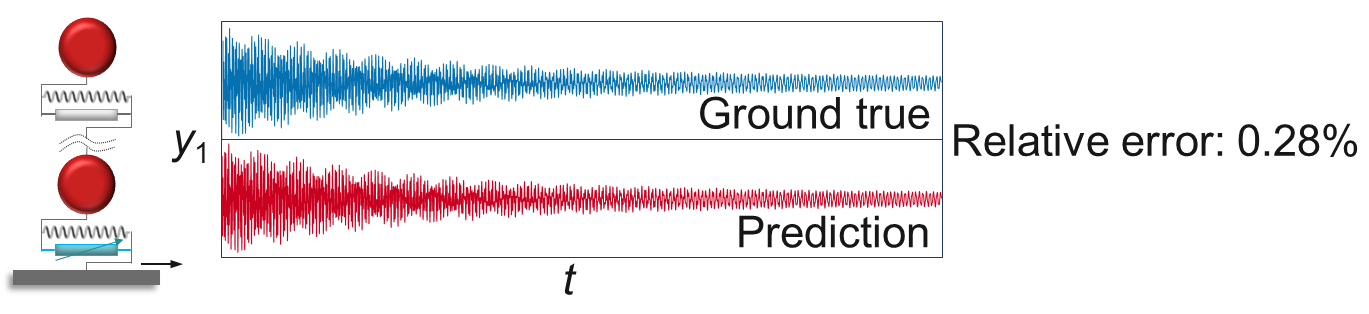} 
} \\

\rowcolor{Myellow!10}
\makecell[c]{Van de Pol}
& \makecell[l]{
$\psi (\mathbf{x}_{k})=({\mathbf{A}_{\upkappa,{0}}}+\mu {\mathbf{A}_{\upkappa,{1}}}+\bar\omega {\mathbf{A}_{\upkappa,{2}}})\psi (\mathbf{x}_{k-1})$ \\
$\mu$ - Nonlinear parameter \\
$\bar\omega$ - Frequency component
}& \makecell[c]{
\includegraphics[width=0.45\textwidth]{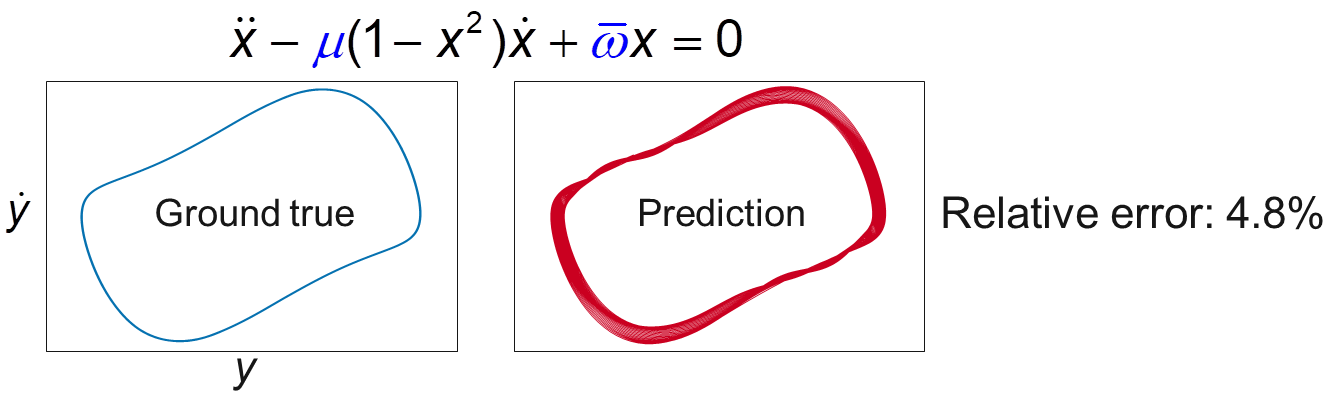} 
} \\

\makecell[c]{Incident-jet}
& \makecell[l]{
${{\mathbf{x}}_{k}}=(\mathbf{A}_0+k_\text{t}{\mathbf{A}_1}){{\mathbf{x}}_{k-1}}$ \\
$k_\text{t}$ - Thermal conductivity
}
& \makecell[c]{
\includegraphics[width = 0.46\textwidth]{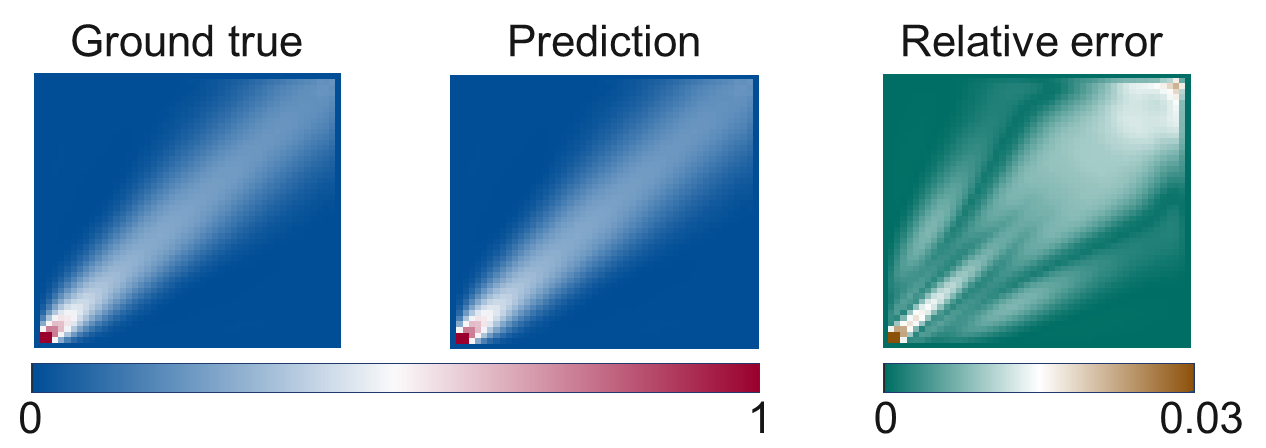}
} \\

\rowcolor{Myellow!10}
\makecell[c]{1-D Burgers'}
& \makecell[l]{
${{\mathbf{x}}_{k}}=(\mathbf{A}_0+v{\mathbf{A}_1}){{\mathbf{x}}_{k-1}}$ \\
$v$ - Viscosity
}
& \makecell[c]{
\includegraphics[width = 0.46\textwidth]{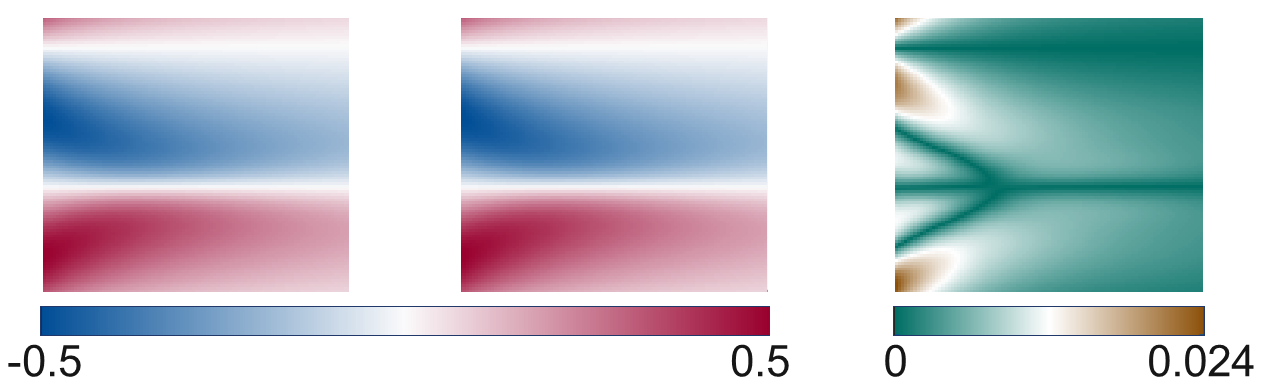}
} \\

\makecell[c]{Cavity flow} 
& \makecell[l]{
${\mathbf{x}_{k}}=(\mathbf{A}_0+v_\text{s}{\mathbf{A}_1}+R_\text{e}{\mathbf{A}_2}){\mathbf{x}_{k-1}}$\\
$v_\text{s}$ - Flow speed \\
$R_\text{e}$ - Reynolds number
}
& \makecell[c]{
\includegraphics[width = 0.46\textwidth]{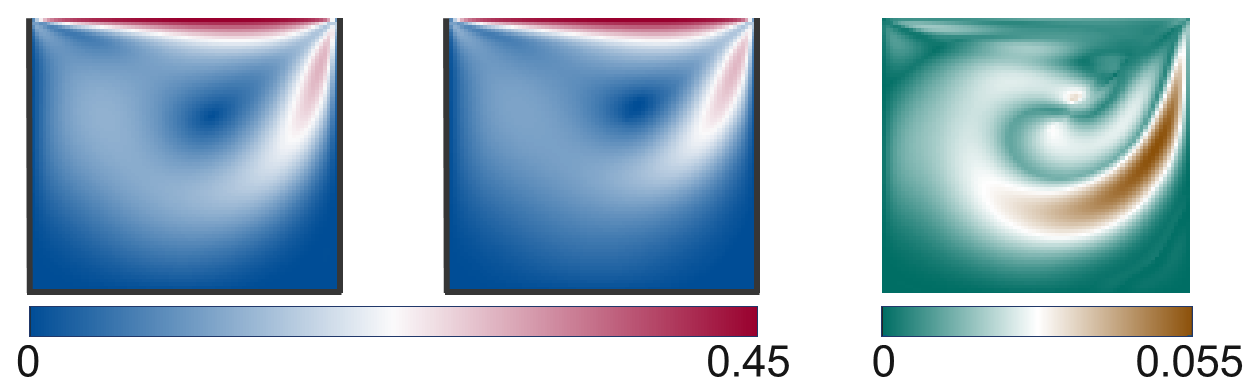}
} \\

\rowcolor{Myellow!10}
\makecell[c]{Smoke plume}
& \makecell[l]{
${\mathbf{x}_{k}}=(\mathbf{A}_0+r_\text{d}{\mathbf{A}_1}){\mathbf{x}_{k-1}}$ \\
$r_\text{d}$ - Radius of the light
}
& \makecell[c]{
\includegraphics[width = 0.46\textwidth]{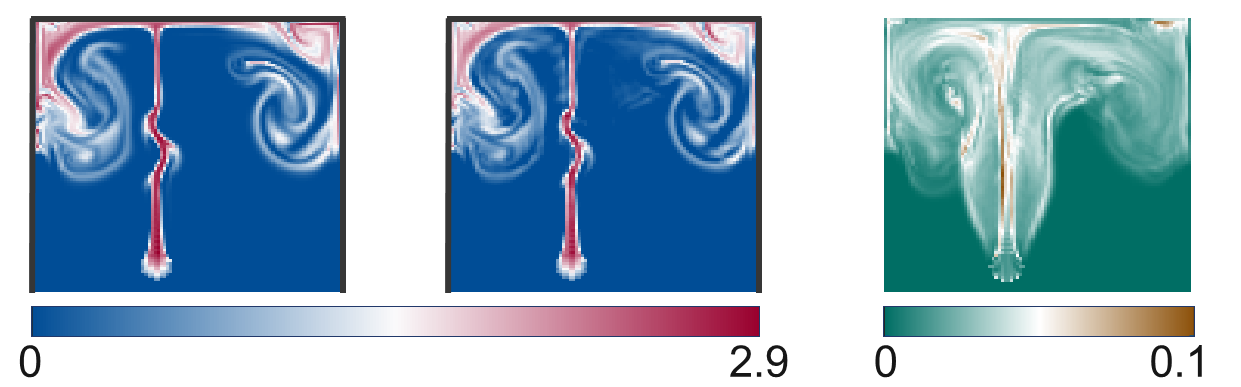}
} \\

\makecell[c]{Droplet control}& \makecell[l]{
${\mathbf{x}_{k}}=(\mathbf{A}_0+{{V}_{\text{t}}}{\mathbf{A}_1}){\mathbf{x}_{k-1}}$ \\
$V_\text{t}$ - Driving voltage
}
& \makecell[c]{
\includegraphics[width = 0.46\textwidth]{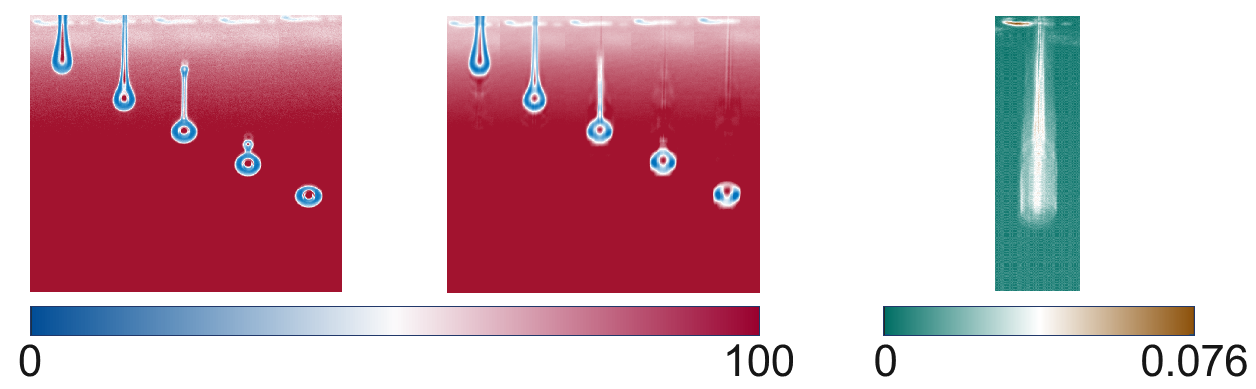}
} \\

\hline
\end{tabular}
      
    \end{minipage}
  \end{table*}
\linespread{1}

\subsection{Extrapolation Capacity and Efficiency}

The proposed ID-DMD method allows for extrapolation across both time and design parameter spaces. This property demonstrates the fundamental difference between the ID-DMD and other machine learning approaches. In this section, we discuss the ID-DMD representation of 1-D Burgers’ equation, comparing it with various operator learning approaches involving PINNs, NIF, Physics-Informed DeepONet (PI-DON), and FNO. All these approaches are extended to enable the parametric simulation of dynamic systems as illustrated in \textcolor{blue}{Fig.S8} and \textcolor{blue}{Tab.8}.

The 1-D Burgers’ equation is defined as~\cite{wang2021learning}:

\begin{equation}
\left\{ \begin{aligned}
  & \frac{\partial s}{\partial t}+s\frac{\partial s}{\partial x}-v\frac{{{\partial}^{2}}s}{\partial {x^{2}}}=0 \\ 
 & s(x,0)=u(x) \\ 
\end{aligned} \right.,\ x\in [0,1] \label{eq5}
\end{equation}

\noindent with periodic boundary conditions

\begin{equation}
s(0,t)=s(1,t)\  \text{and}\  \frac{\partial s(0,t)}{\partial x}=\frac{\partial s(1,t)}{\partial x} \label{eq6}
\end{equation}

\noindent where the initial condition $u(x)$ is generated from a Gaussian random field $u\sim N(0,{{25}^{2}}{{(-\Delta +{{5}^{2}}\mathbf{I})}^{-2}})$ satisfying the periodic boundary conditions. Here, $\Delta$ is the Laplacian operator, and $\mathbf{I}$ is the identity matrix.

\begin{figure*}[!t]
      \centering
      \includegraphics[width=\textwidth]{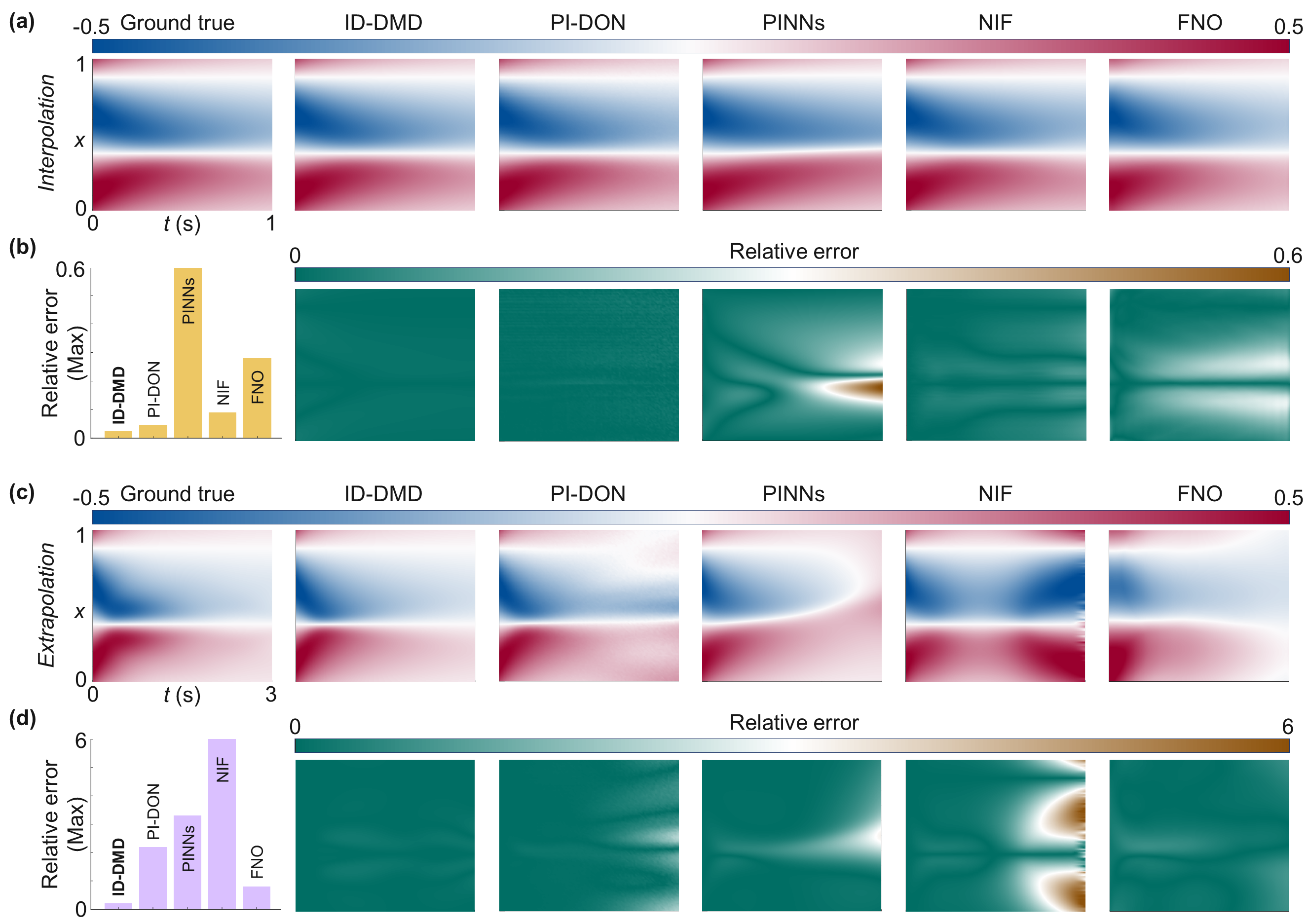}
      \caption{
       Interpolation and extrapolation of the 1-D Burgers’ equation using the ID-DMD, PI-DON, PINNs, NIF, and FNO.  (a) Interpolation prediction results for $v=0.02$ and $t\in [0,1]\text{s}$ using different advanced data-driven methods. (b) Relative errors for the interpolation results. (c) Extrapolation prediction results for $v=0.01$ and $t\in [0,3]\text{s}$ using different advanced data-driven methods. (d) Relative errors for the extrapolation results. 
      }
      \label{fig_2}
\end{figure*}

Fig.\ref{fig_2}  presents a comprehensive comparison of prediction performance across five advanced methods: ID-DMD, PI-DON, PINNs, NIF, and FNO. All models are trained under the same initial conditions, spanning $v\in \left[ 0.014,0.046 \right]$ and $t\in [0,1]\text{s}$ , with spatial resolution defined as $x=[0:0.1:1]$. We examine both the interpolation ($v=0.02$ and $t\in [0,1]\text{s}$ ) and extrapolation ($v=0.01$ and $t\in [0,3]\text{s}$)  performances. Notably, the proposed ID-DMD consistently outperforms its counterparts, delivering the most precise prediction results across both tasks. 

In the interpolation task, all five methods demonstrate strong predictive accuracy for the responses of the Burgers' equation. While the PINNs shows the highest maximum relative error among the approaches, its overall prediction performance remains satisfactory and reliable.

For the extrapolation task, it is evident that all network-based methods (PI-DON, PINNs, NIF, and FNO) struggle with long-term predictions. In contrast, the ID-DMD method excels due to its physically meaningful approach to predicting time series data based on dynamic modes. These modes effectively capture the underlying system dynamics, whereas network-based approaches treat time as a static variable, making long-term prediction a challenging extrapolation problem. 

In general, systems that exhibit periodic or damped behavior can be accurately represented by a finite number of modes, enabling ID-DMD to achieve reliable long-term predictions.  Additionally, the linear-in-parameter structure of the ID-DMD model ensures stable extrapolation over design parameters, further highlighting its robustness and effectiveness in studying dynamic systems.

The ID-DMD demonstrates remarkable efficiency compared to other machine learning approaches, owing to its simple model structure and lightweight identification process. Tab.\ref{tab.2} highlights its superior performance in both training and simulation speeds, along with accuracy in interpolation and extrapolation predictions  (See details in \textcolor{blue}{Section 4 in the Supplementary Materials}).

 It is clear that training the parametric 1-D Burgers' equation using operator learning requires at least 6.5 minutes (PINNs) on a GPU (Tesla T4 @ 16GB). In contrast, the ID-DMD achieves an $\sim\times 10^4$ improvement in speed, completing the process in just 0.02 seconds on a CPU (Intel Core i7-1160G7 @ 16GB). Furthermore, once trained, the ID-DMD delivers comparable simulation speed ($\sim 0.0005 \text{s}$) to all other operator learning approaches, along with the highest prediction accuracies in both interpolation and extrapolation tasks.

  \linespread{1.5}
\begin{table*}[!t] 
      \centering    
\noindent
\caption{Comparison with advanced data-driven approaches}
\label{tab.2}

    \begin{tabular}{|p{2.5cm}|p{3.5cm}|p{3.5cm}|p{3.5cm}|p{3.5cm}|}

\hline
\makecell[c] {Methods} & \makecell[c] {Training\\time ($\sim$)} & \makecell[c]{Simulation\\time ($\sim$)} & \makecell[c]{Max relative error \\(Interpolation)}& \makecell[c]{Max relative error \\ (Extrapolation)}\\

\hline
\makecell[c] {ID-DMD} & \makecell[c] {\textbf{0.02s (CPU)}} & \makecell[c]{\textbf{0.0005s (CPU)}} & \makecell[c]{\textbf{0.024}}& \makecell[c]{\textbf{0.21}}\\

\hline
\makecell[c] {PI-DON} & \makecell[c] {10.65min (GPU)} & \makecell[c]{\textbf{0.0005s (CPU)}} & \makecell[c]{0.047}& \makecell[c]{2.2}\\

\hline
\makecell[c] {PINNs} & \makecell[c] {6.5min (GPU)} & \makecell[c]{0.005s (CPU)} & \makecell[c]{0.6}& \makecell[c]{3.3}\\

\hline
\makecell[c] {NIF} & \makecell[c] {16.5min (GPU)}& \makecell[c]{1s (CPU)} & \makecell[c]{0.09}& \makecell[c]{6}\\

\hline
\makecell[c] {FNO} & \makecell[c] {66.6min (GPU)} & \makecell[c]{0.6s (CPU)} & \makecell[c]{0.28}& \makecell[c]{0.81}\\

\hline
\end{tabular}
\end{table*}
\linespread{1}

\subsection{Physical Interpretability}
This section explains how the ID-DMD is capable of representing a complex dynamic system while also being physically interpretable. In contrast to other data-driven models based on existing operator learning and neural networks, which tend to be complex and operate as black boxes, the ID-DMD is inherently transparent. 

For linear dynamic systems represented by ODE or PDEs, the ID-DMD remains the consistent parametric structure with a discretized physical state space model. Eigenvectors and eigenvalues of the linear operator directly represent system dynamics to enable a straightforward design and optimization. For example, the {\em pole placement} of a 4-DoF linear building system is demonstrated in \textcolor{blue}{Fig.S9}, where the bottom linear stiffness is used as a design parameter. In this example, the state vectors $\mathbf{x}_{k}$ are formulated from the building’s responses ${y_1}(k),{y_2}(k),{y_3}(k),{y_4}(k)$, along with their time-delay embeddings ${y_i}(k-1),{y_i}(k-2),...$, where $i=1,2,3,4$. Using the ID-DMD with the settings in \textcolor{blue}{Tab.S9}. The design target is to achieve the desired first resonant frequency of $10\ \text{rad/s}$ by designing the bottom linear stiffness. The design result is the linear stiffness equal to $2.82\times {10^9}\ {\text{N}}/{\text{m}}$ with a standard deviation of $\pm 0.14\times {{10}^{7}}$. The estimated resonant frequencies of all four modes of the designed system show a maximum relative error of 0.17\%.

Here, we focus more on the interpretability of nonlinear systems, where additional steps are required to project the system’s state measurements into a higher-dimensional linear space. This can be achieved by employing the Koopman operator, making it possible to formulate the ID-DMD representation for nonlinear systems as Eq.(\ref{eq3}).

\begin{figure*}[!bt]
    \centering
    \includegraphics[width=\textwidth]{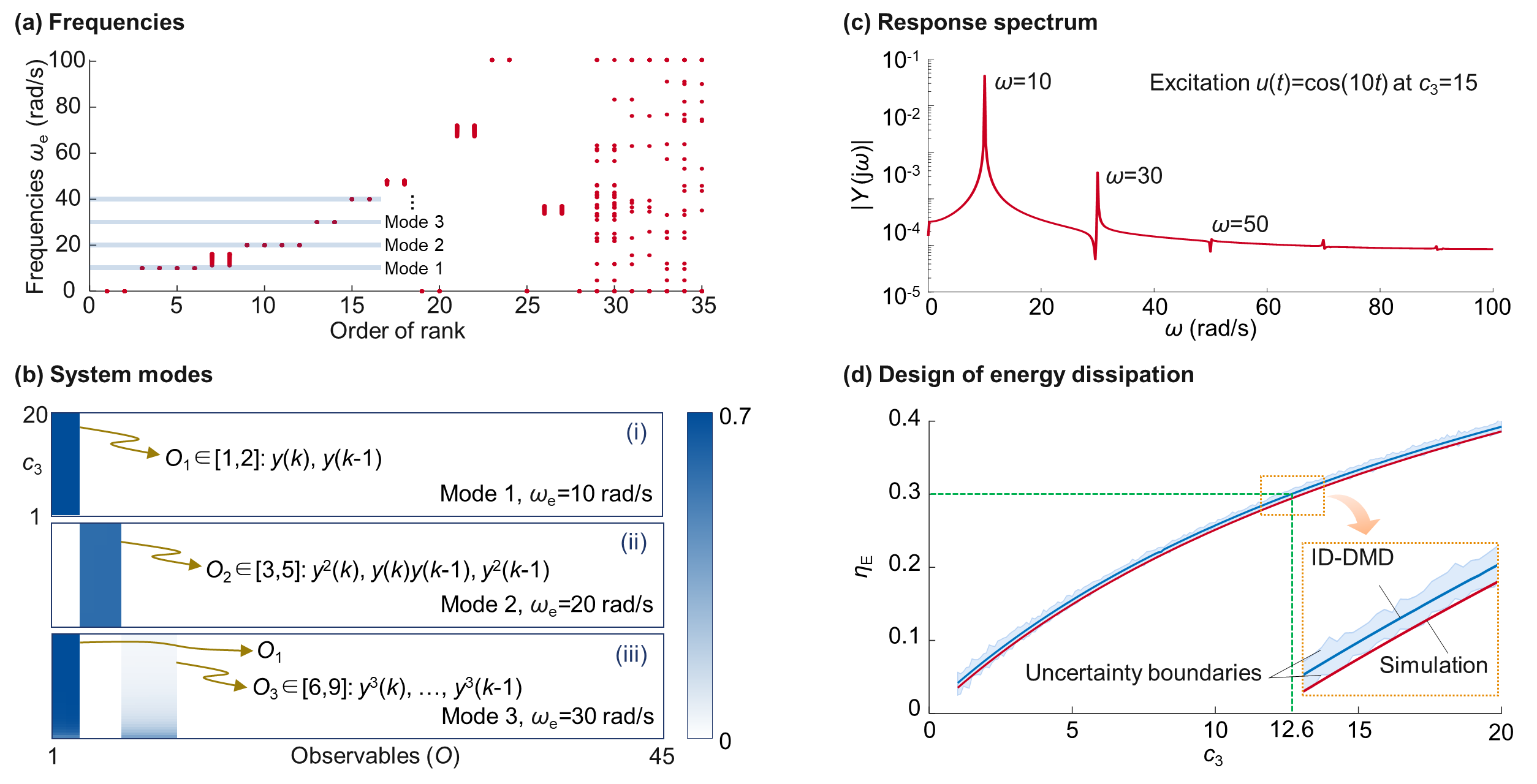}
    \caption{
    Evaluate the first three order modes from the polynomial-projected Koopman operator. (a) The dominant modes and their corresponding characteristic frequencies can be identified from the imaginary part of the eigenvalues in the ID-DMD framework. These dominant modes remain stable across varying design parameters, representing the true dynamics of the system. In contrast, other modes that do not exhibit this stability are classified as spurious modes. (b) The modes of the system with (i) The first mode arises from the linear states, corresponding to the natural frequency of $\omega_\text{e}=10\ \text{rad/s}$. (ii) The second mode results from the squared projection of the states, occurring at the second-order modulation frequency $\omega_\text{e} =20\ \text{rad/s}$. (iii) The third mode, at $\omega_\text{e}=30\ \text{rad/s}$, is contributed by both the linear states and cubic projection of the states. (c) System response spectrum of $y(k)$ under $c_3=15$ and an input excitation $u(t)=\cos(10t)$. (d)  Design the energy dissipation of ${\eta}_{\text{E}}>30\%$ with ${{c}_{3}}>12.6$
    }
    \label{fig_3}
\end{figure*}

In this section, we explore the ID-DMD modeling of a nonlinearly damped ODE:

\begin{equation}
\frac{{{\text{d}}^{2}}y}{\text{d}{{t}^{2}}}+0.03\frac{\text{d}y}{\text{d}t}+100y+{{c}_{3}}(\frac{\text{d}y}{\text{d}t}){}^{3}=0 \label{eq7} 
\end{equation}

\noindent where ${{c}_{3}}$ represents the nonlinear damping and serves as the design parameter. The ID-DMD model of the nonlinearly damped differential equation is $\psi({{\mathbf{x}}_{k}})=({{\mathbf{A}}_{{\upkappa},{0}}}+{{c}_{3}}{\mathbf{A}_{{\upkappa},{1}}})\psi({{\mathbf{x}}_{k-1}})$, where the observables are obtained from the polynomial projection as $\psi({\mathbf{x}}_{k})={[y(k), y(k-1),{{y}^{2}}(k), y(k)y(k-1),{{y}^{2}}(k-1),\ldots ]}^{\text{T}}$ up to the 8th order with in total 45 observables. Settings for ID-DMD identification are shown in \textcolor{blue}{Tab.S10}. Here, Fig.\ref{fig_3} illustrates how the ID-DMD characterizes and designs system dynamics under various nonlinear damping values.

Frequency modulation achieving super- and sub-harmonics is a well-known characteristic of nonlinear systems. Fig.\ref{fig_3}(a) demonstrates the dominant and spurious modes of the ID-DMD.  In Fig.\ref{fig_3}(b), the second mode at $\omega_\text{e}=20\ \text{rad/s}$ does not affect the system's linear state $y(k)$, which means there is no second-order modulation in the system response as shown in Fig.\ref{fig_3}(c). The nonlinear damping increases energy dissipation within the system without affecting the system's settling time~\cite{lang2009theoretical}.  This unique property has been applied in engineering practice, particularly in vibration control and wide-band vibration isolation~\cite{zhu2022beneficial}. The energy dissipation ${{\eta }_{\text{E}}}$ is defined as the energy difference between linear and nonlinear system responses over the energy of the linear response.  A design achieving energy loss greater than $30\%$ is illustrated in Fig.\ref{fig_3}(d). The robustness of the design is evaluated through uncertainty quantification performed using the bagging process. The results shown in Fig.\ref{fig_3} can be readily verified by analytically solving the differential equation Eq.(\ref{eq5}) as detailed in \textcolor{blue}{Section 5 in the Supplementary Materials}.

For more complex nonlinear systems, other projections, i.e., encoding-decoding networks~\cite{takeishi2017learning}, can be applied to implement the ID-DMD using the same method.

\section{Discussion}
ID-DMD is a data-driven, physically interpretable modeling and design framework that can be applied to a broad range of complex dynamic systems at a fraction of the computing cost of competing methods. As parametric modeling is gaining recognition as a promising research direction for system design and optimization, many studies have explored network learning approaches. However, these methods often require the training of complex network structures, and the parametric models themselves tend to function as black boxes. The ID-DMD shows that the dynamic world is not as complex as might be presumed. It demonstrates that parametric data-driven models can be both simple and powerful, offering clear insights into system dynamics. Since ID-DMD can accurately predict system behavior across an underlying parameter space, it has the potential to replace cumbersome numerical simulators, thereby accelerating the analysis and design of dynamic systems. 

This study focuses on the foundational form of ID-DMD, which uses a polynomial extension of states and a linear-in-parameter model structure. While this approach is naive, it encounters limitations when handling discontinuous dynamics and severe nonlinearities. For example, modeling bifurcations, such as those arising from the Kuramoto-Sivashinsky equation, is challenging. Moreover, optimizing structures and geometries across a large nonlinear design space remains difficult with the current ID-DMD approach. However, these issues are less critical in engineering practice, as optimization is typically performed in the neighbour of existing pre-designed configurations~\cite{khuri2010response,schoukens2009robustness}.  In the case of design-from-scratch involving wide-range optimizations, the limitations are expected to be addressed as ID-DMD evolves to a more complete model structure

\begin{equation*}
\psi ({{\mathbf{x}}_{k}})=({{\mathbf{A}}_{{\upkappa},{0}}}+{{f}_{1}}(\bm{\upepsilon}){\mathbf{A}_{{\upkappa},{1}}}+{{f}_{2}}(\bm{\upepsilon}){\mathbf{A}_{{\upkappa},{2}}}+\cdots )\psi({{\mathbf{x}}_{k-1}})
\end{equation*}

\noindent where ${{f}_{i}}(\bm{\upepsilon}),\ i\in {{\mathbb{Z}}^{+}}$ are arbitrary differentiable functions of design parameters, offering a path toward overcoming these challenges and expanding its applicability via auxiliary function learning~\cite{lusch2018deep}. 

\section{Methods}
Existing parametric linear operator identification often requires point-wise regression for high-dimensional matrices~\cite{huhn2023parametric,andreuzzi2023dynamic}. In these studies, each sub-DMD model uses a different projection matrix, resulting in linear operators that are inconsistent and cannot be interpolated. To overcome these issues, the ID-DMD algorithm consists of four key steps: (i) Formulating the regression matrix from the collected snapshots, (ii) Performing dimensionality reduction for ID-DMD, (iii) Predicting system responses, and (iv) Implementing optimal design and uncertainty quantification. Each of these steps is explained in detail below.

\subsection{The regression matrix}
The ID-DMD begins by gathering all the spatio-temporal series data and organizing them into a matrix, denoted as $\mathbf{X}$, where each column represents the states or observables, and the number of columns corresponds to the length of the time series. Under the $l\in {{\mathbb{Z}}^{+}}$th set of design parameter values, the parametric model can be represented in matrix form as:

\begin{equation}
\begin{aligned}
  & {\mathbf{{X}'}}_{(l)}={\mathbf{A}_{0}}{\mathbf{X}_{(l)}}+{\bar{\varepsilon}_{1,(l)}}{\mathbf{A}_{\text{1}}}{\mathbf{X}_{(l)}}+{{\bar{\varepsilon}}_{2,(l)}}{\mathbf{A}_{\text{2}}}{\mathbf{X}_{(l)}}+\cdots \\ 
 & \ \ \ \ =\mathbf{\Theta}{\mathbf{E}_{(l)}} \\ 
\end{aligned} \label{eq8}
\end{equation}

\noindent Here, $\mathbf{X'}=[\begin{matrix}
   {\mathbf{x}_{2}} & {\mathbf{x}_{3}} & \cdots   \\
\end{matrix}]$ is the matrix representing the states at the next time step of $\mathbf{{X}}=[\begin{matrix}
   {\mathbf{x}_{1}} & {\mathbf{x}_{2}} & \cdots   \\
\end{matrix}]$. $\mathbf{\Theta }=[\begin{matrix}
   {{\mathbf{A}}_{0}} & {{\mathbf{A}}_{1}} & {{\mathbf{A}}_{2}} & \cdots \\
\end{matrix}]$, and $\mathbf{E}$ is the regressor matrix that incorporates the design parameters and the state matrix. $\bar{\varepsilon }_{i}={\alpha _{i}}{\varepsilon _{i}},\ i\in {\mathbb{Z}^{+}}$ are scaled design parameters to ensure well-conditioned matrices for calculation, where $\alpha_{i}$ are scaling factors.

The regression matrix for the ID-DMD is formulated by combining all data matrices across different design parameter sets. The newly formed regression matrix is $\mathbf{Z}=\mathbf{\Theta \Xi }$, where $\mathbf{Z}=[\begin{matrix}
   {{{\mathbf{{X}'}}}_{(1)}} & {{{\mathbf{{X}'}}}_{(2)}} & \cdots \\
\end{matrix}]$ and the regressor is $\mathbf{\Xi }=[\begin{matrix}
   {{\mathbf{E}}_{(1)}} & {{\mathbf{E}}_{(2)}} & \cdots \\
\end{matrix}]$.

\subsection{Dimensional reduction }

The core idea of ID-DMD is to find a truncated unitary matrix $\mathbf{U}\in {{\mathbb{C}}^{m\times {{r}_{\text{Z}}}}}$, which reduces the dimensionality of the regression matrix $\mathbf{Z}$ via the SVD $\mathbf{Z }\approx {\mathbf{U}}{\mathbf{\Sigma}}{\mathbf{V}}^{*}$. Here,  “*” is the complex conjugate transpose. $\mathbf{U}$ contains the basis vectors that project the high-dimensional data into a lower-dimensional space, where ${{r}_{\text{Z}}}\le m$, with $m$ being the number of states. Additionally, ID-DMD requires a truncated SVD of the regressor matrix $\mathbf{\Xi }\approx {{\mathbf{U}}_{\Xi}}{{\mathbf{\Sigma}}_{\Xi}}{{\mathbf{V}}_{\Xi}}^{*}$, where the unitary matrix ${{\mathbf{U}}_{\Xi}}$ is truncated with ${{r}_{\Xi}}$ columns.

In this study, ${{r}_{\text{Z}}}$ and ${{r}_{\Xi}}$ are set equal, though they can be adjusted differently as hyper-parameters. These hyper-parameters can be determined by using  The ${{\mathbf{U}}_{\Xi}}$ matrix is divided into several row blocks, represented by ${{\mathbf{U}}_{\Xi,i}}\in {{\mathbb{C}}^{m\times {{r}_{\Xi}}}},\ i\in \mathbb{Z}$, each corresponding to a linear operator ${{\mathbf{A}}_{i}}$ of the system. By projecting the full linear operators into the lower-dimensional through

\begin{equation}
    {{\mathbf{\tilde{A}}}_{i}}={{\mathbf{U}}^{*}}{{\mathbf{A}}_{i}}\mathbf{U}={{\mathbf{U}}^{*}}\mathbf{Z}{{\mathbf{V}}_{\Xi}}{{\mathbf{\Sigma }}_{\Xi}}^{-1}{{\mathbf{U}}_{\Xi,i}}^{*}\mathbf{U} \label{eq9}
\end{equation}

\noindent the eigenvalues and eigenvectors of the system under specific design parameters are evaluated from the reduced operator ${{\mathbf{\tilde{A}}}_{0}}+{{\bar{\varepsilon }}_{1}}{{\mathbf{\tilde{A}}}_{1}}+{{\bar{\varepsilon }}_{2}}{{\mathbf{\tilde{A}}}_{2}}+\cdots$ without working with the full system operator. 

The eigenvalues $\mathbf{\Lambda}\in {{\mathbb{C}}^{{{r}_{\text{Z}}}\times {{r}_{\text{Z}}}}}$ of the reduced operator are retained as those of the full operator. The eigenvectors $\mathbf{\Phi}$, which represent the system’s dynamic modes, are reconstructed from the eigenvectors $\mathbf{W}$ of the reduced operator either via the projected DMD ($\mathbf{\Phi }=\mathbf{UW}$) or the exact DMD~\cite{tu2013dynamic}

\begin{equation*}
    \mathbf{\Phi}=\mathbf{Z}{\mathbf{V}_{\Xi}}{\mathbf{\Sigma}_{\Xi}}^{-1}({\mathbf{U}_{\Xi,0}}^\text{*}+{\bar{\varepsilon}_{1}}{\mathbf{U}_{\Xi,1}}^\text{*}+{\bar{\varepsilon}_{2}}{\mathbf{U}_{\Xi,2}}^\text{*}+\cdots)\mathbf{UW}
\end{equation*}

\subsection{Reconstruction and prediction}

Once the eigenvalues and eigenvectors are determined, we can reconstruct the system states using Eq.(\ref{eq2}), summarizing over $j=1,\ldots,{{r}_{Z}}$. The complex frequency ${{s}_{j}}$ is calculated as ${{s}_{j}}=\Delta {{t}^{-1}}\log ({{\mathbf{\Lambda }}_{(j,j)}})$, where ${{\mathbf{\Lambda }}_{(j,j)}}$ are the diagonal entries of the eigenvalues. In physics, if the complex frequency has a positive real part, the system becomes unstable and diverges. To prevent this, we set ${{\sigma }_{j}}=0$ if ${{\sigma }_{j}}>0$, ensuring stability. This study assumes that the initial system states are known a priori for all predictions. 

\subsection{Design and uncertainty quantification}

One can formulate an optimization problem that the cost function can readily be evaluated from the ID-DMD. Inverse design is done by solving the optimization problem. Uncertainties matrices of the design are assessed by integrating the bagging approach into the ID-DMD~\cite{sashidhar2022bagging}. In this process, random subsets of corresponding columns are selected from the data matrices $\mathbf{X'}$ and $\mathbf{X}$ to construct and implement multiple runs of the ID-DMD. In this study, half of the columns were randomly selected to quantify the design uncertainties. Solving the optimization problem based on these ID-DMD realizations will produce a set of optimal design results. These results are then aggregated and analyzed statistically to estimate the uncertainty matrices, providing a comprehensive evaluation of the reliability of the inverse design process. 

All data, along with the MATLAB and Python codes, are available on GitHub: \url{https://github.com/YZ-Vista/Data-driven-dynamics_ID-DMD}

\vspace{10mm}

\setlength{\parindent}{0pt}

\section*{Acknowledgments}
We acknowledge support from the National Science Foundation AI Institute in Dynamic Systems (grant number 2112085). JNK further acknowledges support from the Air Force Office of Scientific Research (FA9550-19-1-0011). YPZ acknowledges support from the Queen Mary University of London Startup Funding (SEM9307B). ZQL acknowledges support from the UK EPSRC project (EP/R032793/1). BZ acknowledges support from the National Natural Science Foundation of China (1223  2013) and the Natural Science Foundation of Ningxia (2022AAC2003).

\appendix

\onecolumngrid
\section*{Supplementary Materials for:\\
Inverse Design with Dynamic Mode Decomposition}

\section{Section 1: The ID-DMD algorithm}

To identify the ID-DMD of a system, we begin by collecting state data across various design parameters sampled over the design space. These design parameters are typically determined using uniform distributions or other design-of-experiment methods. Since design parameters can span multiple orders of magnitude, it is necessary to ensure well-conditioned matrices for subsequent calculations. To address this issue, we scale the design parameter values as ${\bar{\varepsilon}_{i}}={\alpha_{i}}{\varepsilon_{i}}$, $i\in {\mathbb{Z}^{+}}$, where ${\alpha_{i}}$ are scaling factors.

\subsection{The regression matrix}
The ID-DMD representation is formulated as

\begin{equation}  
    \mathbf{x}_{k}=(\mathbf{A}_{0}+{\bar\varepsilon_{1}}{\mathbf{A}_{1}}+\cdots +{\bar\varepsilon_{P}}{\mathbf{A}_{P}}){\mathbf{x}_{k-1}} \label{eqS1}
\end{equation}

\noindent or in a matrix form

\begin{equation}  
    \mathbf{X'}=(\mathbf{A}_{0}+{\bar\varepsilon_{1}}{\mathbf{A}_{1}}+\cdots +{\bar\varepsilon_{P}}{\mathbf{A}_{P}})\mathbf{X} \label{eqS2}
\end{equation}

\noindent where $\bar\varepsilon_{1},\ldots,\bar\varepsilon_{P}$, $P\in {\mathbb{Z}^{+}}$, are the scaled design parameters; $\mathbf{x}_{k}$ is the state vector or snapshot at the discrete time $k$, \\

\begin{equation*} 
    \mathbf{X'}=\left[\begin{matrix}
       | & | & |  \\
    \mathbf{x}_{2} & \mathbf{x}_{3} & \cdots   \\
       | & | & |  \\
    \end{matrix} \right] \ \text{and} \ \mathbf{X}=\left[\begin{matrix}
       | & | & |  \\
    \mathbf{x}_{1} & \mathbf{x}_{2} & \cdots   \\
       | & | & |  \\
    \end{matrix} \right]
\end{equation*} 

Eq.(\ref{eq2}) can be rewritten as

\begin{equation}
    \mathbf{X'}=\mathbf{\Theta E} \label{eqS3}
\end{equation}

\noindent where 

\begin{equation*}
    \mathbf{\Theta}=[\mathbf{A}_{0}\ \mathbf{A}_{1}\ \cdots \ \mathbf{A}_{P}] \ \text{and} \ \mathbf{E}=\left[\begin{matrix}
    \mathbf{X} \\
    {\bar\varepsilon_{1}}\mathbf{X} \\
    \vdots \\
    {\bar\varepsilon_{P}}\mathbf{X} \\
\end{matrix} \right]
\end{equation*}

For different sets of scaled design parameters $\{\bar\varepsilon_{1},\ldots ,\bar\varepsilon_{P}\}$, the ID-DMD shares the same linear operators $\mathbf{\Theta}$. As a result, a unified matrix representation for the ID-DMD can be expressed as: 

\begin{equation}
\mathbf{Z}=\mathbf{\Theta \Xi} \label{eqS4}
\end{equation}

\noindent where 

\begin{equation*}
    \mathbf{Z}=[{\mathbf{X'}_{(1)}}\ {\mathbf{X'}_{(2)}}\ \cdots]\in{\mathbb{R}^{m\times n}} \ \text{and} \ \mathbf{\Xi}=[{\mathbf{E}_{(1)}}\ {\mathbf{E}}_{(2)}\cdots]\in {\mathbb{R}^{(P+1)m\times n}}, m,n\in {\mathbb{Z}^{+}}
\end{equation*}

\noindent with ${\mathbf{X'}_{(l)}}$ and ${\mathbf{E}_{(l)}}$ being matrices under the $l\in {\mathbb{Z}^{+}}$th set of design parameters.

\subsection{Dimensional reduction}
Conducting the SVD of $\mathbf{Z}$ and $\mathbf{\Xi}$, yields

\begin{equation}
    \mathbf{Z} =\underbrace{\mathbf{\bar{U}}}_{m\times m}\mathbf{\bar\Sigma }\underbrace{\mathbf{\bar{V}}^\text{*}}_{m\times n}\approx \underbrace{\mathbf{U}}_{m\times {r}_{\text{Z}}}\mathbf{\Sigma }\underbrace{\mathbf{V}^\text{*}}_{{{r}_{\text{Z}}}\times n} \label{eqS5}
\end{equation}

\noindent and

\begin{equation}
    \underbrace{\mathbf{\Xi}}_{(P+1)m\times n}=\underbrace{\mathbf{\bar{U}}_{\Xi}}_{(P+1)m\times (P+1)m}{\mathbf{\bar\Sigma}_{\Xi}}{{\underbrace{\mathbf{\bar{V}}_{\Xi}}_{n\times n}}^\text{*}}\approx \underbrace{\mathbf{U}_{\Xi}}_{(P+1)m\times {r_{\Xi}}}{\mathbf{\Sigma}_{\Xi}}\underbrace{{\mathbf{V}_{\Xi}}^\text{*}}_{{{r}_{\Xi}}\times n} \label{eqS6}
\end{equation}

\noindent Here, “*” is the complex conjugate transpose. The truncated ranks ${r}_{\text{Z}}$ and ${r}_{\Xi}$ are hyper-parameters used in the identification of the ID-DMD. Their values, constrained by ${r}_{\text{Z}},{r}_{\Xi}\le m$, can be determined by using various techniques for hard and soft thresholding of singular values~\cite{gavish2014optimal,brunton2022data}. The matrix $\mathbf{U}_{\Xi}$ can be further organized as a block matrix:

\begin{equation*}
    \mathbf{U}_{\Xi}=\left[\begin{matrix}
    \underbrace{\mathbf{U}_{\Xi,0}}_{m\times {{r}_{\Xi}}} \\
    \underbrace{\mathbf{U}_{\Xi,1}}_{m\times {{r}_{\Xi}}} \\
    \vdots \\
    \underbrace{\mathbf{U}_{\Xi,P}}_{m\times {{r}_{\Xi}}} \\
    \end{matrix} \right]\in {\mathbb{C}^{(P+1)m\times {{r}_{\Xi}}}}
\end{equation*}

From Eq.(\ref{eqS4}), the linear operators can be evaluated as

\begin{equation}
    \underbrace{\mathbf{\Theta}}_{m\times (P+1)m}=\underbrace{\mathbf{Z}}_{m\times n}\underbrace{\mathbf{\Xi}^{\dagger}}_{n\times (P+1)m}=\underbrace{\mathbf{Z}}_{m\times n}\underbrace{\mathbf{V}_{\Xi}}_{n\times {{r}_{\Xi}}}{\mathbf{\Sigma }_{\Xi}}^{-1}\underbrace{{\mathbf{U}_{\Xi}}^{\text{*}}}_{{{r}_{\Xi}}\times (P+1)m}=\underbrace{\mathbf{Z}}_{m\times n}\underbrace{\mathbf{V}_{\Xi}}_{n\times {{r}_{\Xi}}}{{\mathbf{\Sigma }}_{\Xi}}^{-1}[\underbrace{{\mathbf{U}_{\Xi,0}}^\text{*}}_{{{r}_{\Xi}}\times m}\ \underbrace{{\mathbf{U}_{\Xi,1}}^\text{*}}_{{{r}_{\Xi}}\times m}\ \cdots \ \underbrace{{\mathbf{U}_{\Xi,P}}^\text{*}}_{{{r}_{\Xi}}\times m}] \label{eqS7}
\end{equation}

\noindent where in $\mathbf{\Theta}=[\mathbf{A}_{0}\ \mathbf{A}_{1}\ \cdots \ \mathbf{A}_{P}]$,

\begin{equation*}
    \mathbf{A}_{i}=\mathbf{Z}{\mathbf{V}_{\Xi}}{{\mathbf{\Sigma }}_{\Xi}}^{-1}{\mathbf{U}_{\Xi,i}}^\text{*},\ i\in \mathbb{Z}
\end{equation*}

Next, two Propositions will be proposed for the low-rank representation of ID-DMD. 

\textbf{Proposition 1:} The low-rank ID-DMD is obtained as

\begin{equation}
    \mathbf{\tilde{X}'}=({\mathbf{\tilde{A}}_{0}}+{\bar{\varepsilon}_{1}}{\mathbf{\tilde{A}}}_{1}+\cdots+{\bar{\varepsilon}_{P}}{\mathbf{\tilde{A}}_{P}})\mathbf{\tilde{X}} \label{eqS8}
\end{equation}

\noindent where $\mathbf{X}=\mathbf{U\tilde{X}}$, and

\begin{equation*}    
    \mathbf{\tilde{A}}_{i} =\underbrace{\mathbf{U}^\text{*}}_{{{r}_{\text{Z}}}\times m}{\mathbf{A}_{i}}\underbrace{\mathbf{U}}_{m\times {{r}_{\text{Z}}}}={\mathbf{U}^\text{*}}\mathbf{Z}{\mathbf{V}_{\Xi}}{\mathbf{\Sigma }_{\Xi}}^{-1}{\mathbf{U}_{\Xi,i}}^\text{*}\mathbf{U},\ i\in \mathbb{Z}
\end{equation*}

\textbf{Proof 1:} The ID-DMD can be projected to a low-rank space through the eigenvectors $\mathbf{U}$ as

\begin{equation}
    \begin{aligned}
    & \mathbf{X'}=({\mathbf{A}_{0}}+{{\bar{\varepsilon}}_{1}}{\mathbf{A}_{1}}+\cdots+{{\bar{\varepsilon }}_{P}}{\mathbf{A}_{P}})\mathbf{X} \\ 
    & \Rightarrow \mathbf{U\tilde{X}'}=({\mathbf{A}_{0}}+{{\bar{\varepsilon }}_{1}}{\mathbf{A}_{1}}+\cdots+{{\bar{\varepsilon}}_{P}}{\mathbf{A}_{P}})\mathbf{U\tilde{X}} \\ 
    & \Rightarrow {\mathbf{U}^{*}}\mathbf{U\tilde{X}'}={\mathbf{U}^{*}}({\mathbf{A}_{0}}+{{\bar{\varepsilon }}_{1}}{\mathbf{A}_{1}}+\cdots+{{\bar{\varepsilon}}_{P}}{\mathbf{A}_{P}})\mathbf{U\tilde{X}} \\ 
    & \Rightarrow \mathbf{\tilde{X}'}=({\mathbf{U}^{*}}{\mathbf{A}_{0}}\mathbf{U}+{{\bar{\varepsilon}}_{1}}{\mathbf{U}^{*}}{\mathbf{A}_{1}}\mathbf{U}+\cdots+{{\bar{\varepsilon}}_{P}}{\mathbf{U}^{*}}{\mathbf{A}_{P}}\mathbf{U})\mathbf{\tilde{X}} \\ 
    & \Rightarrow \mathbf{\tilde{X}'}=({{\mathbf{\tilde{A}}}_{0}}+{{\bar{\varepsilon }}_{1}}{{\mathbf{\tilde{A}}}_{1}}+\cdots+{{\bar{\varepsilon}}_{P}}{{\mathbf{\tilde{A}}}_{P}})\mathbf{\tilde{X}} \\ 
    \end{aligned} \label{eqS9}
\end{equation}

Here, it’s worth noting that ${\mathbf{U}^\text{*}}\mathbf{U}=\mathbf{I}$ while $\mathbf{U}{\mathbf{U}^\text{*}}\ne \mathbf{I}$.

\textbf{Proposition 2:} The eigenvalues of $\mathbf{\tilde{A}}_{0}+{\bar{\varepsilon}_{1}}{\mathbf{\tilde{A}}_{1}}+\cdots+{\bar{\varepsilon}_{P}}{\mathbf{\tilde{A}}_{P}}$ are the same as those of ${\mathbf{A}_{0}}+{\bar{\varepsilon}_{1}}{\mathbf{A}_{1}}+\cdots+{\bar{\varepsilon}_{P}}{\mathbf{A}_{P}}$. The eigenvectors of the full-rank operator $\mathbf{\Phi}$ are reconstructed by either the exact DMD method~\cite{tu2013dynamic,brunton2022data}:

\begin{equation}
    \mathbf{\Phi}=\mathbf{Z}{\mathbf{V}_{\Xi}}{\mathbf{\Sigma}_{\Xi}}^{-1}({\mathbf{U}_{\Xi,0}}^\text{*}+{\bar{\varepsilon}_{1}}{\mathbf{U}_{\Xi,1}}^\text{*}+\cdots+{\bar{\varepsilon}_{P}}{\mathbf{U}_{\Xi,P}}^\text{*})\mathbf{UW} \label{eqS10}
\end{equation}

\noindent or the projected DMD method ~\cite{schmid2010dynamic}

\begin{equation}
    \mathbf{\Phi}=\mathbf{UW} \label{eqS11}
\end{equation}

\noindent where $\mathbf{W}$ is the eigenvector of the low-rank operator $\mathbf{\tilde{A}}_{0}+{\bar{\varepsilon}_{1}}{\mathbf{\tilde{A}}_{1}}+\cdots+{\bar{\varepsilon}_{p}}{\mathbf{\tilde{A}}_{p}}$.

\textbf{Proof 2:} Considering the columns of $\mathbf{U}$ span the column space $\mathbf{Z}$, there is~\cite{tu2013dynamic}

\begin{equation}
    {\mathbf{U}^\text{*}}\mathbf{Z}\approx {\mathbf{U}^\text{*}}\mathbf{U\Sigma}{\mathbf{V}^\text{*}}=\mathbf{\Sigma}{\mathbf{V}^\text{*}}\Rightarrow \mathbf{U}{\mathbf{U}^\text{*}}\mathbf{Z}\approx \mathbf{U\Sigma}{\mathbf{V}^\text{*}}=\mathbf{Z} \label{eqS12}
\end{equation}

The eigenvalue decomposition of the low-rank operator can be expressed as $(\mathbf{\tilde{A}}_{0}+{\bar{\varepsilon}_{1}}{\mathbf{\tilde{A}}_{1}}+\cdots+{\bar{\varepsilon}_{P}}{\mathbf{\tilde{A}}_{P}})\mathbf{W}=\mathbf{W\Lambda}$, where $\mathbf{\Lambda}$ contains the eigenvalues, and $\mathbf{W}$ is the matrix of corresponding eigenvectors. 

Now, define the matrix $\mathbf{\Phi}$ (the exact DMD) as:

\begin{equation*}
    \mathbf{\Phi}=\mathbf{Z}{\mathbf{V}_{\Xi}}{\mathbf{\Sigma}_{\Xi}}^{-1}({\mathbf{U}_{\Xi,0}}^\text{*}+{\bar{\varepsilon}_{1}}{\mathbf{U}_{\Xi,1}}^\text{*}+\cdots+{\bar{\varepsilon}_{P}}{\mathbf{U}_{\Xi,P}}^\text{*})\mathbf{UW}
\end{equation*}

\noindent there is 

\begin{equation}
    \begin{aligned}
    & (\mathbf{A}_{0}+{\bar{\varepsilon}_{1}}{\mathbf{A}_{1}}+\cdots+{\bar{\varepsilon}_{P}}{\mathbf{A}_{P}})\mathbf{\Phi}=[\mathbf{Z}{\mathbf{V}_{\Xi}}{\mathbf{\Sigma}_{\Xi}}^{-1}({\mathbf{U}_{\Xi,0}}^\text{*}+{\bar{\varepsilon}_{1}}{\mathbf{U}_{\Xi,1}}^\text{*}+\cdots+{\bar{\varepsilon}_{P}}{\mathbf{U}_{\Xi,P}}^\text{*})]\times  \\ 
    & [\mathbf{Z}{\mathbf{V}_{\Xi}}{\mathbf{\Sigma}_{\Xi}}^{-1}({\mathbf{U}_{\Xi,0}}^\text{*}+{\bar{\varepsilon}_{1}}{\mathbf{U}_{\Xi,1}}^\text{*}+\cdots+{\bar{\varepsilon}_{P}}{\mathbf{U}_{\Xi,P}}^\text{*})\mathbf{UW}] \\ 
    & \approx [\mathbf{Z}{\mathbf{V}_{\Xi}}{\mathbf{\Sigma}_{\Xi}}^{-1}({\mathbf{U}_{\Xi,0}}^\text{*}+{\bar{\varepsilon}_{1}}{\mathbf{U}_{\Xi,1}}^\text{*}+\cdots+{\bar{\varepsilon}_{P}}{\mathbf{U}_{\Xi,P}}^\text{*})]\times \\ 
    & [\mathbf{U}\underbrace{{\mathbf{U}^\text{*}}\mathbf{Z}{\mathbf{V}_{\Xi}}{\mathbf{\Sigma}_{\Xi}}^{-1}({\mathbf{U}_{\Xi,0}}^\text{*}+{\bar{\varepsilon}_{1}}{\mathbf{U}_{\Xi,1}}^\text{*}+\cdots+{\bar{\varepsilon}_{P}}{\mathbf{U}_{\Xi,P}}^\text{*})\mathbf{U}}_{{\mathbf{U}^\text{*}}({\mathbf{A}_{0}}+{\bar{\varepsilon}_{1}}{\mathbf{A}_{1}}+\cdots+{\bar{\varepsilon}_{P}}{\mathbf{A}_{P}})\mathbf{U}=\mathbf{\tilde{A}}_{0}+{\bar{\varepsilon}_{1}}{\mathbf{\tilde{A}}_{1}}+\cdots+{\bar{\varepsilon}_{P}}{{\mathbf{\tilde{A}}}_{P}}}\mathbf{W}] \\ 
    & =\underbrace{[\mathbf{Z}{\mathbf{V}_{\Xi}}{\mathbf{\Sigma}_{\Xi}}^{-1}({\mathbf{U}_{\Xi,0}}^\text{*}+{\bar{\varepsilon}_{1}}{\mathbf{U}_{\Xi,1}}^\text{*}+\cdots+{\bar{\varepsilon}_{P}}{\mathbf{U}_{\Xi,P}}^\text{*})][\mathbf{UW}}_{\mathbf{\Phi}}\mathbf{\Lambda}]=\mathbf{\Phi \Lambda} \\ 
    \end{aligned} \label{eqS13}
\end{equation}

On the other hand, denote $\mathbf{\Phi }=\mathbf{UW}$ (the projected DMD), since $\mathbf{U}{\mathbf{U}^\text{*}}\mathbf{Z}\approx \mathbf{Z}$, there is

\begin{equation*}
    \begin{aligned}
    & \mathbf{A}_{0}+{\bar{\varepsilon}_{1}}{\mathbf{A}_{1}}+\cdots +{\bar{\varepsilon}_{P}}{\mathbf{A}_{P}} \\ 
    & =\mathbf{Z}{\mathbf{V}_{\Xi}}{\mathbf{\Sigma}_{\Xi}}^{-1}{\mathbf{U}_{\Xi,0}}^\text{*}+{\bar{\varepsilon}_{1}}\mathbf{Z}{\mathbf{V}_{\Xi}}{\mathbf{\Sigma}_{\Xi}}^{-1}{\mathbf{U}_{\Xi,1}}^\text{*}+\cdots +{\bar{\varepsilon}_{P}}\mathbf{Z}{\mathbf{V}_{\Xi}}{\mathbf{\Sigma}_{\Xi}}^{-1}{\mathbf{U}_{\Xi,P}}^\text{*} \\ 
    & \approx (\mathbf{U}{\mathbf{U}^\text{*}}\mathbf{Z}{\mathbf{V}_{\Xi}}{\mathbf{\Sigma}_{\Xi}}^{-1}{\mathbf{U}_{\Xi,0}}^\text{*}+{\bar{\varepsilon}_{1}}\mathbf{U}{\mathbf{U}^\text{*}}\mathbf{Z}{\mathbf{V}_{\Xi}}{\mathbf{\Sigma}_{\Xi}}^{-1}{\mathbf{U}_{\Xi,1}}^\text{*}+\cdots +{\bar{\varepsilon}_{P}}\mathbf{U}{\mathbf{U}^\text{*}}\mathbf{Z}{\mathbf{V}_{\Xi}}{\mathbf{\Sigma}_{\Xi}}^{-1}{\mathbf{U}_{\Xi,P}}^\text{*}) \\ 
    & =\mathbf{U}{\mathbf{U}^\text{*}}({\mathbf{A}_{0}}+{\bar{\varepsilon}_{1}}{\mathbf{A}_{1}}+\cdots+{\bar{\varepsilon}_{P}}{\mathbf{A}_{P}}) \\ 
    \end{aligned}
\end{equation*}

Then we have

\begin{equation}
    \begin{aligned}
    & ({\mathbf{A}_{0}}+{\bar{\varepsilon}_{1}}{\mathbf{A}_{1}}+\cdots+{\bar{\varepsilon}_{P}}{\mathbf{A}_{P}})\mathbf{\Phi}=({\mathbf{A}_{0}}+{\bar{\varepsilon}_{1}}{\mathbf{A}}_{1}+\cdots+{\bar{\varepsilon}_{P}}{\mathbf{A}_{P}})\mathbf{UW} \\ 
    & =\mathbf{U}\underbrace{{\mathbf{U}^\text{*}}({\mathbf{A}_{0}}+{\bar{\varepsilon}_{1}}{\mathbf{A}_{1}}+\cdots+{\bar{\varepsilon}_{P}}{\mathbf{A}_{P}})\mathbf{U}}_{\mathbf{\tilde{A}_{0}}+{\bar{\varepsilon}_{1}}{\mathbf{\tilde{A}}_{1}}+\cdots+{\bar{\varepsilon}_{P}}{\mathbf{\tilde{A}}_{P}}}\mathbf{W}=\mathbf{UW\Lambda}=\mathbf{\Phi \Lambda} \\ 
    \end{aligned} \label{eqS14}
\end{equation}

\subsection{Reconstruction and prediction}
The predicted output response is

\begin{equation}
    \mathbf{x}_{k}=\sum\limits_{j\in \mathbb{Z}^{+}}{{\bm{\upphi}_{j}}{\text{e}^{{{s}_{j}}(k-1)}}{{b}_{j}}}=\mathbf{\Phi}\exp [\mathbf{S}(k-1)]\mathbf{b} \label{eqS16}
\end{equation}

\noindent where $\mathbf{b}={\mathbf{\Phi}^{\dagger}}{\mathbf{x}_{1}}$, ${\mathbf{x}_{1}}$ represents the initial states of the system; $\mathbf{S}$ is a diagonal matrix containing the elements ${s}_{j}={\sigma_{j}}+\text{j}{\omega_{j}}$ with ${\omega_{j}}$ being the characteristic frequency and ${\sigma_{j}}$ being the decay rate is the continuous time eigenvalues, evaluated as 

\begin{equation}
    {s}_{j}=\Delta {t}^{-1}\log ({\mathbf{\Lambda}_{(j,j)}}) \label{eqS15}
\end{equation}

\noindent where $\mathbf{\Lambda}_{(j,j)}$ are the diagonal entries of the eigenvalues, $\Delta t$ is the sampling time.

In physics, if the complex frequency has a positive real part, the system becomes unstable and diverges. To prevent this, we set ${\sigma_{j}}=0$ if ${\sigma_{j}}>0$, ensuring stability.

For model validation, the relative error~\cite{tofallis2015better} at each discrete time step $k$ is defined as

\begin{equation}
    \bm{\upeta}_{k}=\frac{\left| \mathbf{x}_{k,\text{Prediction}}-\mathbf{x}_{k,\text{True}} \right|}{\max (\left| \mathbf{x}_{k,\text{True}} \right|)}\times 100\%  \label{eqS17}
\end{equation}

\noindent and the total relative error (applied to Tab.I) is defined as the mean values of $\bm{\upeta}_{k}$ over time. The relative error is an effective measure for evaluating errors of varying amplitudes over time. 


\section{Section 2: The pitched airfoil}
\begin{figure}[!htb]
  \centering
  \includegraphics[width=0.45\textwidth]{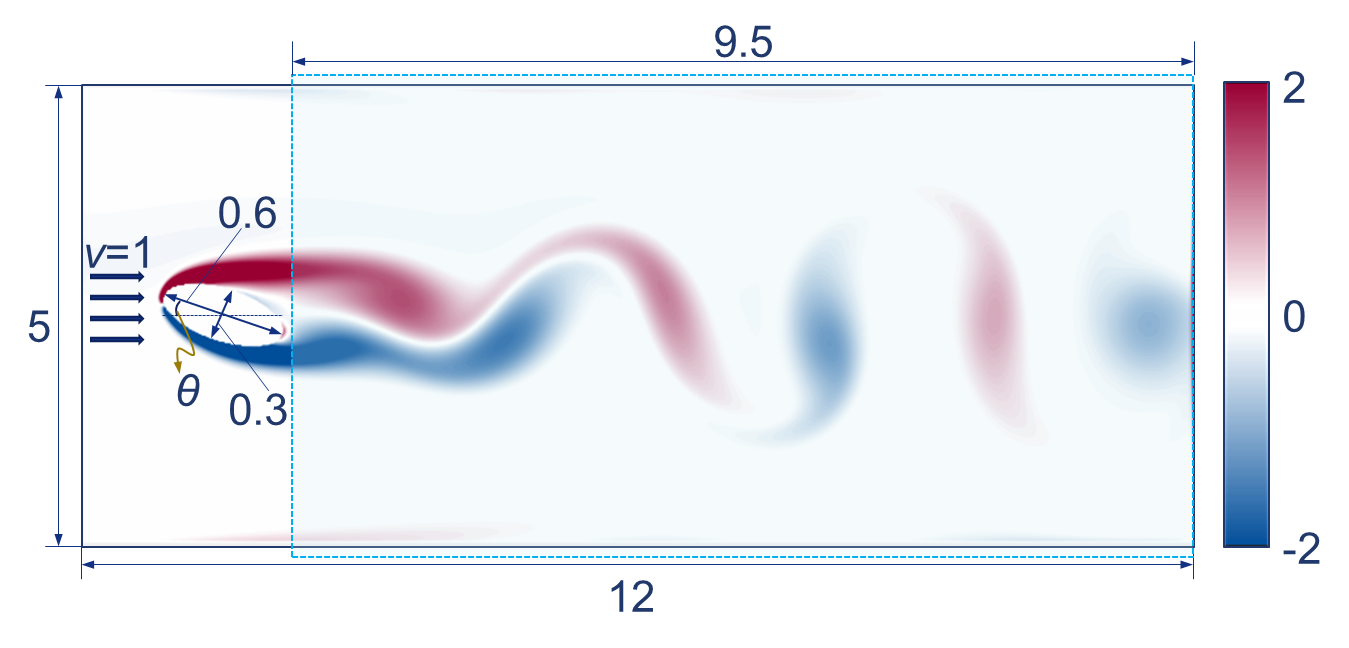}
  \caption{
 Configuration for the vorticity simulation of a pitched airfoil.
  }
  \label{S1}
\end{figure}

The pitched airfoil under investigation is simplified to an elliptical shape with a dimensionless length of 0.6 and a width of 0.3. The inlet airflow speed is set to 1, corresponding to a Reynolds number of 150. The simulation is performed using the Lattice Boltzmann Method~\cite{aidun2010lattice}. The configuration and simulation setup of the pitched airfoil are illustrated in Fig.\ref{S1}.

In this example, random noise with a zero mean and a standard deviation of 0.3 (approximately 15\% noise) is added to the measurements (Fig.1(a)). The resulting ID-DMD for the pitched airfoil is (Fig.1(b))

\begin{equation}
    \mathbf{x}_{k}=(\mathbf{A}_{0}+\theta {\mathbf{A}_{1}}){\mathbf{x}_{k-1}} \label{eqS18}
\end{equation}

\noindent where the steady-state training data are across $\theta =\{2, 4, 6, 8, 10, 12\}{}^\circ$ over $t\in [20,50]\ \text{s}$ with the sampling time $\Delta t=0.5\ \text{s}$.

\begin{figure}[!htb]
  \centering
  \includegraphics[width=0.75\textwidth]{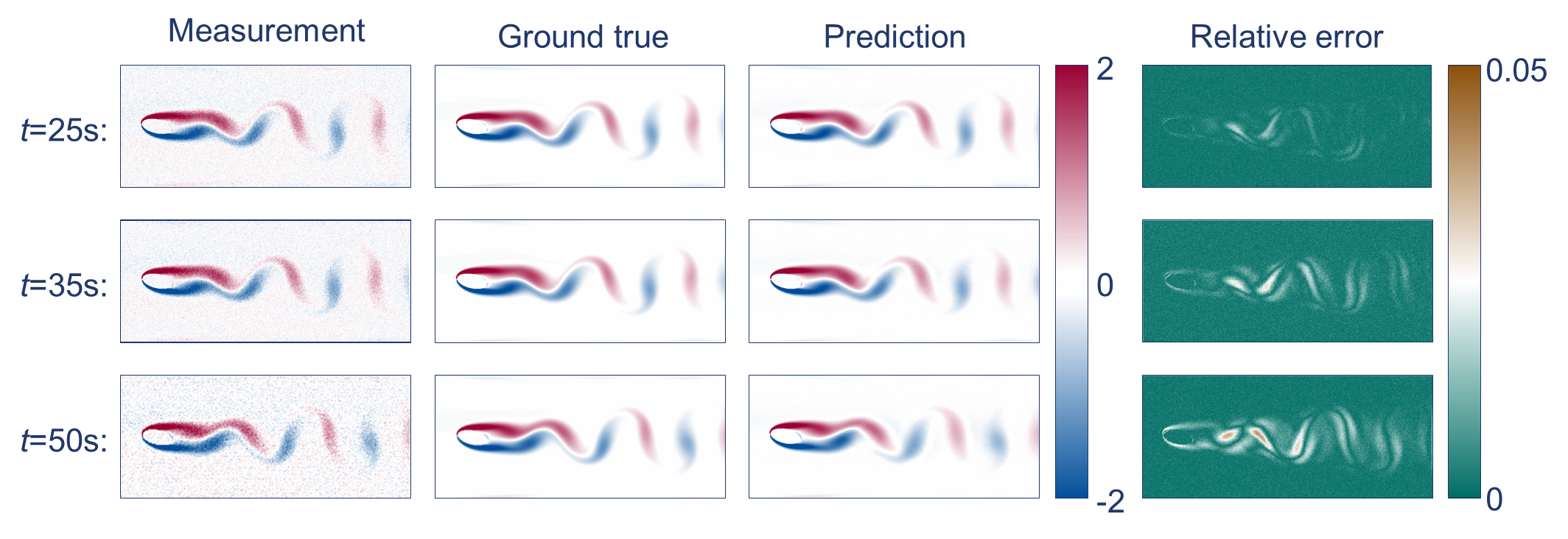}
  \caption{
  Prediction of the vorticity distribution around the airfoil.
  }
  \label{S2}
\end{figure}

For the ID-DMD identification, we set the hyper-parameters ${r}_{\text{Z}}={r}_{\Xi}=120$, with a scaling factor of $\alpha =1$, to achieve optimal modeling and fitting performance. The ID-DMD predictions of the vorticity at, i.e. $\theta =5^\circ$, are shown in Fig.1(c) and detailed results are shown in Fig.\ref{S2}.

The design objective is to minimize the vorticity power corresponding to a specific wavelength. This objective is formulated as the following optimization problem (Fig.1(d)): 

\begin{equation}
    \theta =\arg \min ({P}_\text{air})\ \ \text{s}\text{.t}\text{.}\ \left\{\begin{aligned}
    & \lambda \ge 3.35 \\ 
    & \theta \in {\mathbb{Z}^{+}} \\ 
    \end{aligned} \right. \label{eqS19}
\end{equation}

Here, the power ${P}_\text{air}$ is defined as

\begin{equation}
    {P}_\text{air}=\frac{1}{N}\sqrt{\sum\limits_{x,y,k}{{\mathbf{x}_{k}}^{2}}} \label{eqS20}
\end{equation}

\noindent where $x,y$ are coordinates of the flow field over the blue dotted box in Fig.\ref{S1}. $N$ is the number of snapshots.

In addition, the uncertainties in the power predictions are quantified using the bagging method. In this process, we randomly select half the columns ($n/2$) of $\mathbf{Z}$ and $\mathbf{\Xi}$ to evaluate the eigenvalues and eigenvectors. Following the identification procedure of the ID-DMD, we repeat this process 30 times, and the statistical results of the power ${P}_\text{air}$ are shown in Fig.1(e). The optimal pitched angle is designed as $\theta =7^\circ$.

\section{Section 3: ID-DMD representation of complex dynamic systems}

In this section, we provide a detailed explanation of all the examples presented in Tab.I. The 1-D Burgers' equation will be discussed separately in Section IV to facilitate a comparison of advanced machine learning methods.

\subsection{Nonlinearly damped building}
\begin{figure}[!h]
  \centering
  \includegraphics[width=0.65\textwidth]{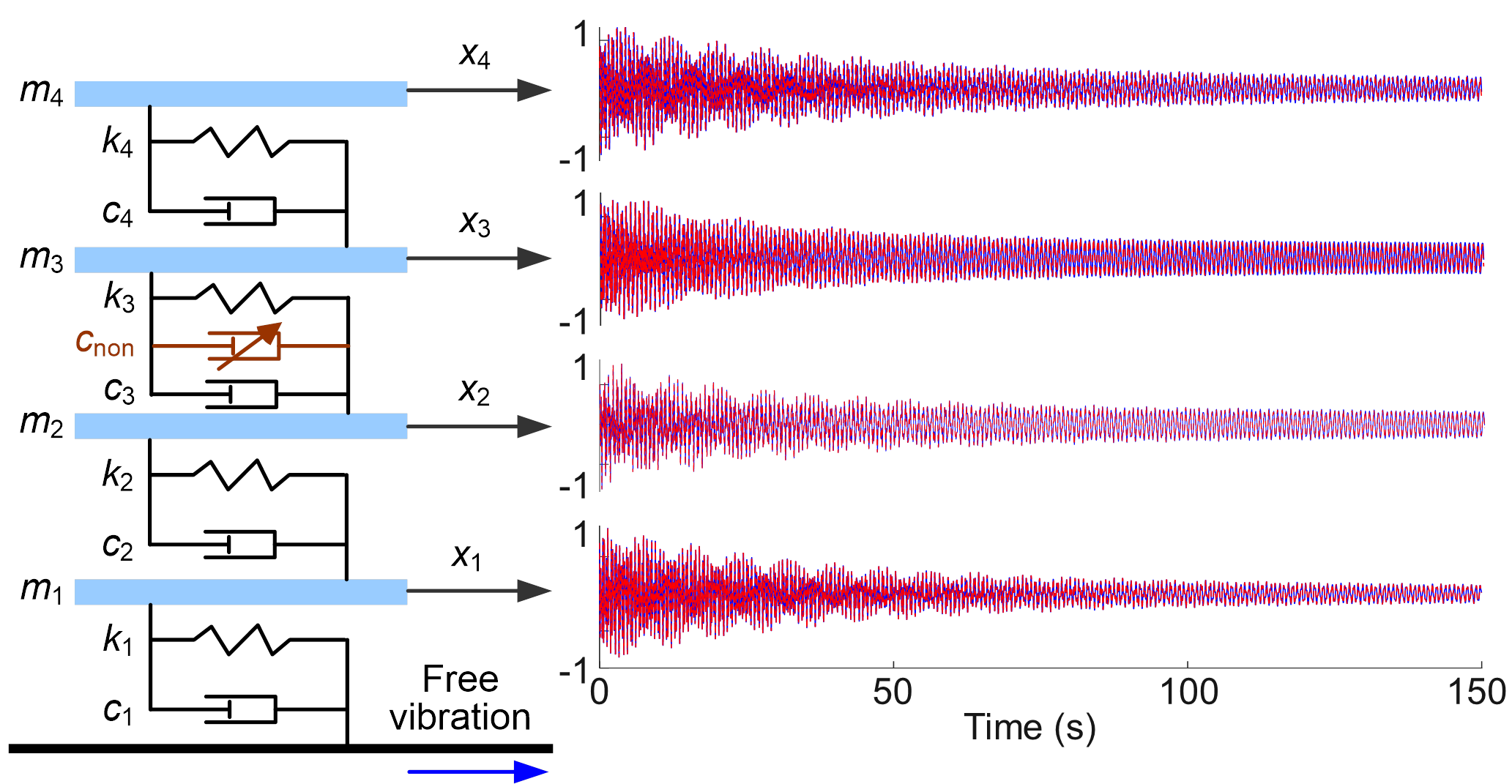}
  \caption{
  Nonlinearly damped building structure and the ID-DMD prediction results.
  }
  \label{S3}
\end{figure}

Consider a nonlinearly damped 4-Degree-of-Freedom (4-DoF) building system as illustrated in Fig.\ref{S3}~\cite{zhu2022design}. This system is governed by the ODE as 

\begin{equation}
    \mathbf{M\ddot{x}}+ \mathbf{C\dot{x}}+\mathbf{Kx}+{{\mathbf{F}}_{\text{non}}}=\mathbf{0}  \label{eqS21}
\end{equation}

\noindent with

\begin{equation*}
    \mathbf{x}=\left[\begin{matrix}
    {{x}_{1}}(t)  \\
    {{x}_{2}}(t)  \\
    {{x}_{3}}(t)  \\
    {{x}_{4}}(t)  \\
    \end{matrix} \right],\ \mathbf{M}=\left[\begin{matrix}
    {m}_{1} & 0 & 0 & 0  \\
    0 & {m}_{2} & 0 & 0  \\
    0 & 0 & {m}_{3} & 0  \\
    0 & 0 & 0 & {m}_{4}  \\
    \end{matrix} \right],\ \mathbf{C}=\left[\begin{matrix}
    {{c}_{1}}+{{c}_{2}} & -{c}_{2} & 0 & 0  \\
    -{c}_{2} & {{c}_{2}}+{{c}_{3}} & -{c}_{3} & 0  \\
    0 & -{c}_{3} & {{c}_{3}}+{{c}_{4}} & -{c}_{4}  \\
    0 & 0 & -{c}_{4} & {c}_{4}  \\
    \end{matrix} \right]
\end{equation*}

\begin{equation*}
    \mathbf{K}=\left[ \begin{matrix}
    {{k}_{1}}+{{k}_{2}} & -{k}_{2} & 0 & 0  \\
    -{k}_{2} & {{k}_{2}}+{{k}_{3}} & -{k}_{3} & 0  \\
    0 & -{k}_{3} & {{k}_{3}}+{{k}_{4}} & -{k}_{4}  \\
    0 & 0 & -{k}_{4} & {k}_{4}  \\
    \end{matrix} \right],\ \text{and} \ {\mathbf{F}_\text{non}}=\left[ \begin{matrix}
    0  \\
    -{{c}_{\text{non}}}{{({\dot{x}_{3}}-{\dot{x}_{2}})}^{3}}  \\
    {{c}_\text{non}}{{({\dot{x}_{3}}-{\dot{x}_{2}})}^{3}}  \\
    0  \\
    \end{matrix} \right]
\end{equation*}

\noindent where the masses (Kg) of the four floors are defined as ${m}_{1}=5\times {{10}^{6}}$, ${m}_{2}=4\times {{10}^{6}}$, ${m}_{3}=3\times {{10}^{6}}$ and ${m}_{4}=2\times {{10}^{6}}$; The stiffness (N/m) of the building’s inter-floor connections are ${k}_{1}=1500\times {{10}^{6}}$, ${k}_{2}=2000\times {{10}^{6}}$, ${k}_{3}=3000\times {{10}^{6}}$ and ${k}_{4}=1000\times {{10}^{6}}$; The linear damping (Nm/s) for all stories are uniform and set to ${c}_{1}={c}_{2}={c}_{3}={c}_{4}=1\times {{10}^{5}}$. The building undergoes free vibration with initial displacements of ${{x}_{1}}(0)=1\ \text{m}$ and ${{x}_{2}}(0)={{x}_{3}}(0)={{x}_{4}}(0)=0$ and the initial velocities being all zero. The nonlinear damper ${{c}_\text{non}}\in [100,5000]\ \text{N}{\text{m}^{3}}/{\text{s}^{3}}$  placed between the masses ${m}_{2}$ and ${m}_{3}$ is the design parameter. 

To capture the system dynamics, a polynomial Koopman operator is applied. This expansion generates a set of augmented states with time delay, which is constructed to represent the dynamics of the system in a higher-dimensional space, allowing effective modeling and analysis of nonlinear behavior. The expanded states include polynomial terms up to the 8th degree:

\begin{equation*}
    \psi ({\mathbf{x}_{k}})={{[{{x}_{1}}(k),{{x}_{1}}(k-1),{{x}_{2}}(k),\ldots,{{x}_{4}}(k-1),{{x}_{2}}(k){{x}_{3}}(k),\ldots,{{x}_{2}}(k-1){{x}_{3}}^{7}(k),...]}^\text{T}}
\end{equation*}

The settings used for ID-DMD identification are listed in Tab.\ref{tab.S1}. The predicted responses at ${c}_\text{non}=500$ are shown in Fig.\ref{S3}.

\linespread{1.2}
\begin{table*}[!h] 
    \centering    
\noindent
\caption{ID-DMD settings for the nonlinearly damped building}
\label{tab.S1}

    \begin{tabular}{|p{4cm}|p{9cm}|}

\hline
\makecell[l] {Training parameter} & \makecell[l] {${{c}_{\text{non}}}=\{\text{0}\text{.1, 0}\text{.8, 3, 5}\}\times {{10}^{3}}$} \\

\hline
\makecell[l] {Time period} & \makecell[l] {$t\in [0,150]\ \text{s}$} \\

\hline
\makecell[l] {Sampling time} & \makecell[l] {$\Delta t=1/64\ \text{s}$} \\

\hline
\makecell[l] {Hyper-parameters} & \makecell[l] {${r}_\text{Z}={r}_{\Xi}=100$} \\

\hline
\makecell[l] {Scaling factor for ${c}_\text{non}$} & \makecell[l] {$\alpha =0.001$} \\

\hline
\makecell[l] {ID-DMD} & \makecell[l] {$\psi (\mathbf{x}_{k})=({\mathbf{A}_{\kappa,0}}+{{c}_\text{non}}{\mathbf{A}_{\kappa,1}})\psi (\mathbf{x}_{k-1})$} \\

\hline
\end{tabular}
\end{table*}
\linespread{1}

\subsection{Van De Pol equation}
The Van De Pol equation is 

\begin{equation}
    \ddot{x}-\mu (1-{x}^{2})\dot{x}+\bar{\omega}x=0 \label{eqS22}
\end{equation}

\noindent with the initial condition $x(0)=0.1$ and $\dot{x}(0)=0$. The variables $\mu \in [0.8,1.2]$ and $\bar{\omega}\in [0.8,1.2]$ serve as the two design parameters in this analysis. Here, a polynomial Koopman operator up to an order of 11 with 77 observables is applied to capture the steady states of the Van De Pol equation: 

\begin{equation*}
    \psi (\mathbf{x}_{k})={[x(k),x(k-1),{{x}^{2}}(k),\ldots ,{{x}^{11}}(k-1)]^\text{T}}
\end{equation*}

The settings used for ID-DMD identification are listed in Tab.\ref{tab.S2}. The predicted responses at $(\mu,\bar{\omega})=(1,1.1)$ are shown in Tab.I. 

\linespread{1.2}
\begin{table*}[!ht] 
    \centering    
\noindent
\caption{ID-DMD settings for the Van De Pol equation}
\label{tab.S2}

    \begin{tabular}{|p{4cm}|p{9cm}|}

\hline
\makecell[l] {Training parameter} & \makecell[l] {$\begin{aligned}
  & (\mu,\bar{\omega})=\{\text{(0}\text{.8,0}\text{.8), (0}\text{.8,1), (0}\text{.8,1}\text{.2),}\ \text{(0}\text{.9,0}\text{.9),}\,\text{(0}\text{.9,1}\text{.1),}\ \text{(1,0}\text{.8),} \\ 
 & \text{(1,1),}\ \text{(1,1}\text{.2),}\ \text{(1}\text{.1,0}\text{.9),}\ \text{(1}\text{.1,1}\text{.1),}\ \text{(1}\text{.2,0}\text{.8),}\ \text{(1}\text{.2,1),}\ \text{(1}\text{.2,1}\text{.2)}\} \\ 
\end{aligned}$} \\

\hline
\makecell[l] {Time period} & \makecell[l] {$t\in [30,200]\ \text{s}$} \\

\hline
\makecell[l] {Sampling time} & \makecell[l] {$\Delta t=1/32\ \text{s}$} \\

\hline
\makecell[l] {Hyper-parameters} & \makecell[l] {${r}_\text{Z}={r}_{\Xi}=70$} \\

\hline
\makecell[l] {Scaling factor for $(\mu,\bar{\omega})$} & \makecell[l] {$(\alpha_{1},\ \alpha_{2})=(1,\ 1)$} \\

\hline
\makecell[l] {ID-DMD} & \makecell[l] {$\psi (\mathbf{x}_{k})=(\mathbf{A}_{\kappa,0}+\mu {\mathbf{A}_{\kappa,1}}+\bar{\omega}{\mathbf{A}_{\kappa,2}})\psi (\mathbf{x}_{k-1})$} \\

\hline
\end{tabular}
\end{table*}
\linespread{1}

\subsection{Incident-jet flow}
\begin{figure}[!ht]
  \centering
  \includegraphics[width=0.65\textwidth]{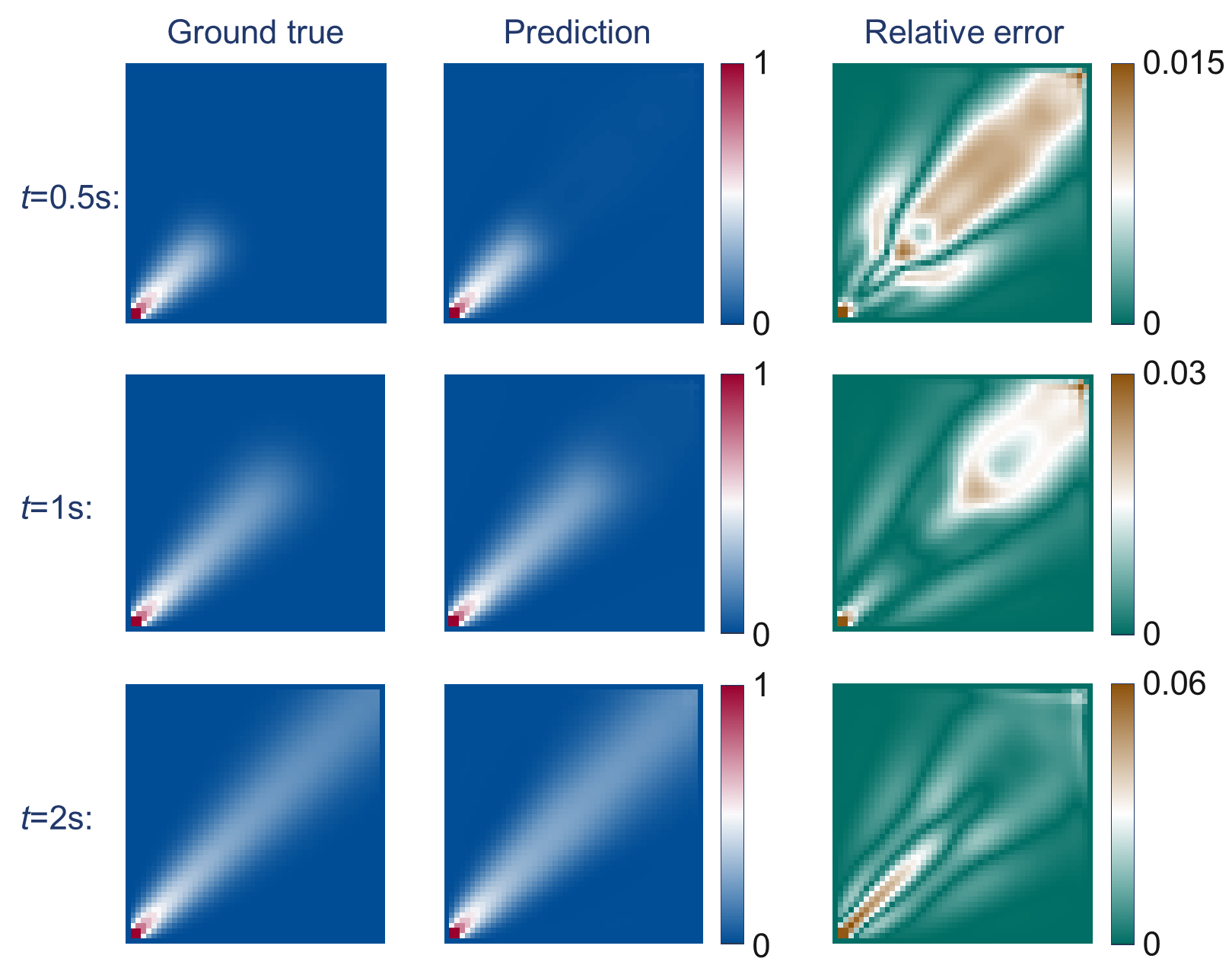}
  \caption{
  Prediction of the incident-jet flow.
  }
  \label{S4}
\end{figure}

The incident-jet flow~\cite{huhn2023parametric} is represented by the governing equation

\begin{equation}
    \frac{\partial T}{\partial t}+w\nabla T=\nabla \cdot {{k}_{\text{t}}}\nabla T \label{eqS23}
\end{equation}

\noindent where the simulation domain is defined as $\{(x,y)\in [0,5]\otimes [0,5]\}$. At the boundary, the temperature is set as $T(x<0.3,y<0.3,t)=1$. The advection velocity is $w=3$, and the thermal conductivity ${{k}_\text{t}}\in [0.2,0.8]$ is the design parameter. The settings for ID-DMD identification are provided in Tab.\ref{tab.S3}. The prediction results for ${k}_\text{t}=0.3$ are illustrated in Fig.\ref{S4}.

\linespread{1.2}
\begin{table*}[!ht] 
    \centering    
\noindent
\caption{ID-DMD settings for the incident-jet flow}
\label{tab.S3}

    \begin{tabular}{|p{4cm}|p{9cm}|}

\hline
\makecell[l] {Training parameter} & \makecell[l] {${k}_\text{t}=\{0.2,0.4,0.6,0.8\}$} \\

\hline
\makecell[l] {Time period} & \makecell[l] {$t\in [0,2]\ \text{s}$} \\

\hline
\makecell[l] {Sampling time} & \makecell[l] {$\Delta t=0.025\ \text{s}$} \\

\hline
\makecell[l] {Hyper-parameters} & \makecell[l] {${r}_\text{Z}={r}_{\Xi}=150$} \\

\hline
\makecell[l] {Scaling factor for ${k}_\text{t}$} & \makecell[l] {$\alpha =1$} \\

\hline
\makecell[l] {ID-DMD} & \makecell[l] {$\mathbf{x}_{k}=(\mathbf{A}_{0}+{{k}_\text{t}}{\mathbf{A}_{1}}){\mathbf{x}_{k-1}}$} \\

\hline
\end{tabular}
\end{table*}
\linespread{1}

\subsection{Cavity flow}
The PDE of the Cavity flow is represented by a Navier-Stokes equation~\cite{bruneau20062d}:

\begin{equation}
    \left\{\begin{aligned}
    & \frac{\partial \mathbf{u}}{\partial t}+(\mathbf{u}\cdot \nabla)\mathbf{u}=-\nabla p+\frac{1}{{R}_\text{e}}{{\nabla}^{2}}\mathbf{u} \\ 
    & \nabla \mathbf{u}=0 \\ 
    \end{aligned} \right. \label{eqS24}
\end{equation}

\noindent where $p$ is the pressure. $\mathbf{u}(x,y,t)$ is the velocity field with $x,y\in [0,1]$. ${R}_\text{e}$ is the Reynold number. Here, the boundary conditions are $\mathbf{u}(x,y,t)=(0,0)$ at the left, bottom and right walls and $\mathbf{u}(x,y,t)=(v,0)$ on top, where $v$ is the driven velocity of the flow. The initial condition is $\mathbf{u}(x,y,0)=(0,0)$ as the flow starts from rest. 

\begin{figure}[!ht]
  \centering
  \includegraphics[width=0.65\textwidth]{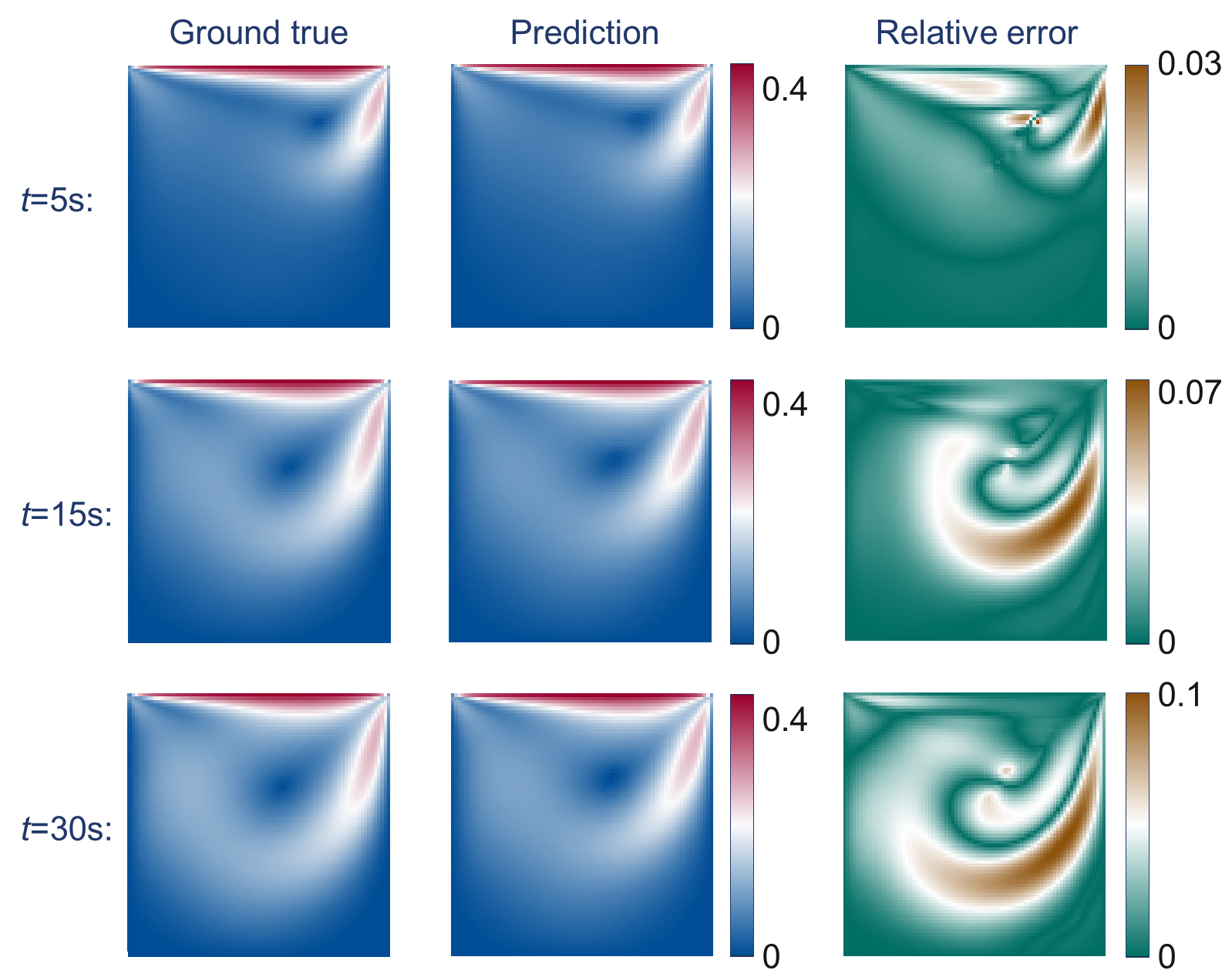}
  \caption{
  Prediction of the cavity flow.
  }
  \label{S5}
\end{figure}

The cavity flow design parameters are ${R}_\text{e}\in [400,800]$ and $v\in [0.4,0.8]$. The settings for ID-DMD identification are listed in Tab.\ref{tab.S4}. The ID-DMD test results for cavity flow at $({R}_\text{e},\ v)=(500,0.5)$ are shown in Fig.\ref{S5}.

\linespread{1.2}
\begin{table*}[!ht] 
    \centering    
\noindent
\caption{ID-DMD settings for the Cavity flow}
\label{tab.S4}

    \begin{tabular}{|p{4cm}|p{9cm}|}

\hline
\makecell[l] {Training parameter} & \makecell[l] {$\begin{aligned}
  & (v,{R}_\text{e})=\{(0.4,400), (0.4,800), (0.5,700), (0.6,600),
  \\ 
 & (0.7,500), (0.8,400), (0.8,800)\} \\ 
\end{aligned}$} \\

\hline
\makecell[l] {Time period} & \makecell[l] {$t\in [0,30]\ \text{s}$} \\

\hline
\makecell[l] {Sampling time} & \makecell[l] {$\Delta t=0.2\ \text{s}$} \\

\hline
\makecell[l] {Hyper-parameters} & \makecell[l] {${r}_\text{Z}={r}_{\Xi}=60$} \\

\hline
\makecell[l] {Scaling factor for $(v, {{R}_{\text{e}}})$} & \makecell[l] {$(\alpha_{1},\alpha_{2})=(1,0.001)$} \\

\hline
\makecell[l] {ID-DMD} & \makecell[l] {$\mathbf{x}_{k}=(\mathbf{A}_{0}+v{\mathbf{A}_{1}}+{{R}_\text{e}}{\mathbf{A}_{2}}){\mathbf{x}_{k-1}}$} \\

\hline
\end{tabular}
\end{table*}
\linespread{1}

\subsection{Smoke plume}
The smoke plume data used in this study is generated using the Python code available at

\url{https://github.com/Ceyron/machine-learning-and-simulation/tree/main/english/phiflow}. 

\begin{figure}[!ht]
  \centering
  \includegraphics[width=0.65\textwidth]{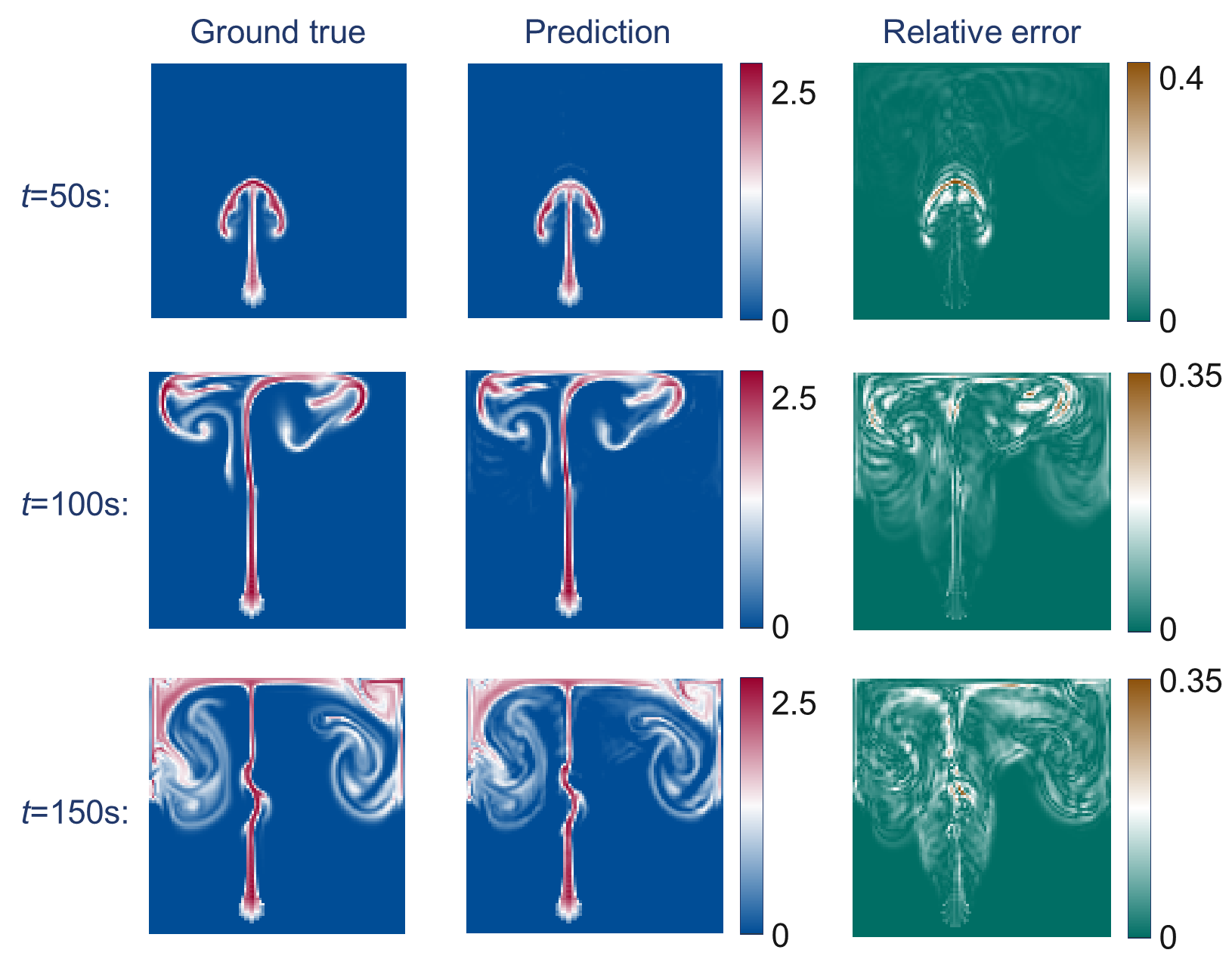}
  \caption{
  Prediction of the smoke plume.
  }
  \label{S6}
\end{figure}

The primary design parameter is the dimensionless radius of the burner, denoted as ${r}_\text{d}$. Simulations are conducted for a range of ${{r}_{\text{d}}}\in [5,5.5]$, representing variations in burner size that influence the smoke plume dynamics. The simulation captures the intricate flow patterns associated with the smoke plume, providing a rich dataset for analysis. The ID-DMD identification method is applied to this dataset to model the system's dynamics, with detailed settings provided in Tab.\ref{tab.S5}. The testing results of the ID-DMD for the smoke plume at ${r}_\text{d}=5.2$ are shown in Fig.\ref{S6}.

\linespread{1.2}
\begin{table*}[!ht] 
    \centering    
\noindent
\caption{ID-DMD settings for smoke plume}
\label{tab.S5}

    \begin{tabular}{|p{4cm}|p{9cm}|}

\hline
\makecell[l] {Training parameter} & \makecell[l] {${r}_\text{d}=\left\{ 5, 5.1, 5.4, 5.5 \right\}$} \\

\hline
\makecell[l] {Time period} & \makecell[l] {$t\in [0,150]\ \text{s}$} \\

\hline
\makecell[l] {Sampling time} & \makecell[l] {$\Delta t=1\ \text{s}$} \\

\hline
\makecell[l] {Hyper-parameters} & \makecell[l] {${r}_\text{Z}={r}_{\Xi}=350$} \\

\hline
\makecell[l] {Scaling factor for ${r}_\text{d}$} & \makecell[l] {$\alpha=0.1$} \\

\hline
\makecell[l] {ID-DMD} & \makecell[l] {$\mathbf{x}_{k}=(\mathbf{A}_{0}+{{r}_\text{d}}{\mathbf{A}_{1}}){\mathbf{x}_{k-1}}$} \\

\hline
\end{tabular}
\end{table*}
\linespread{1}

\subsection{Droplet}
\begin{figure}[!ht]
  \centering
  \includegraphics[width=0.7\textwidth]{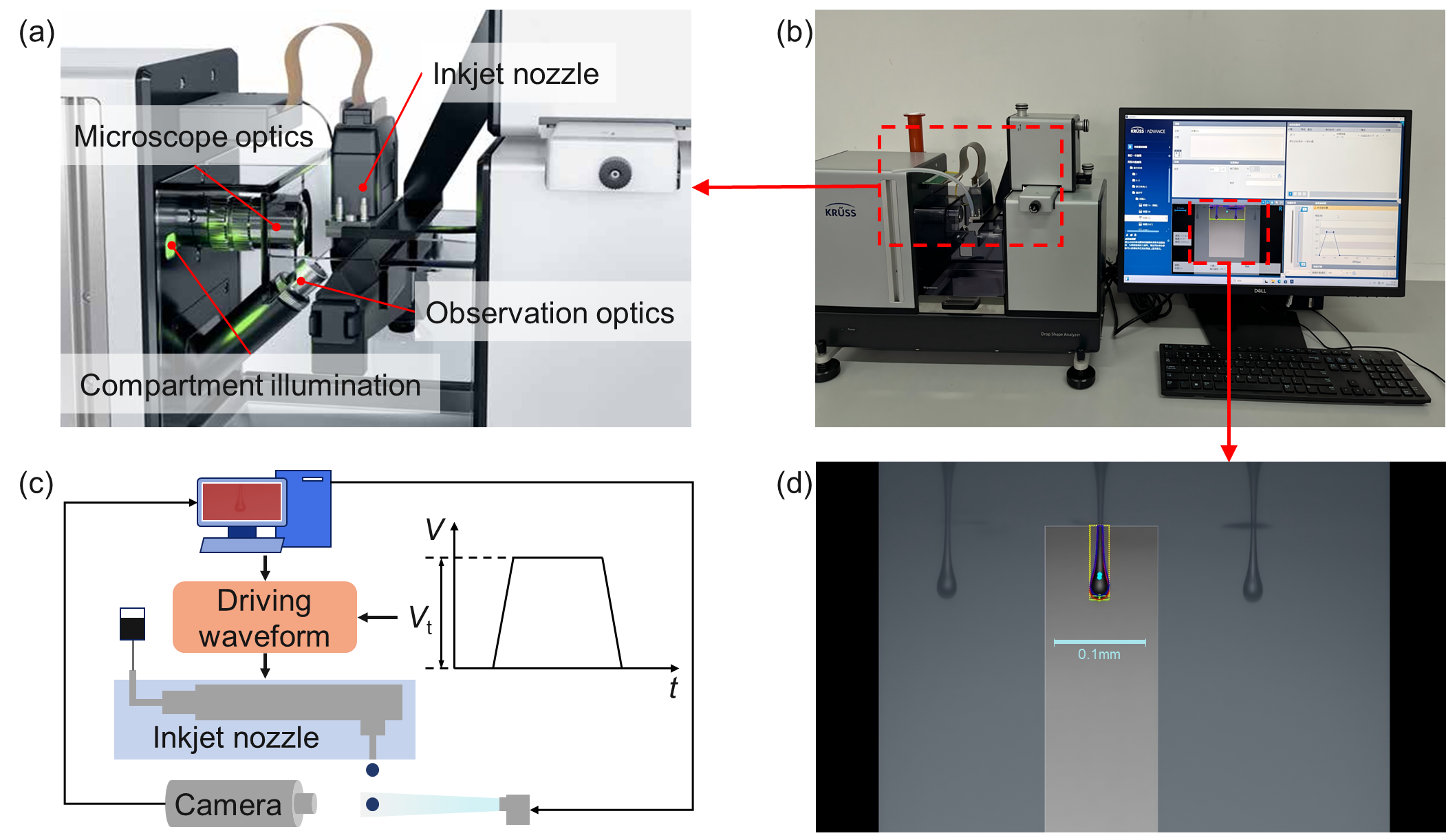}
  \caption{
  Experimental setup for the droplet test. (a) Overview of the main components of the DSA Inkjet system. (b) Illustration of the experimental setup during operation. (c) Configuration of the ink control and supply system. (d) Droplet observation during the experiment.
  }
  \label{S7}
\end{figure}

The inkjet nozzle is the core component of the droplet-based 3D printing process. The accurate implementation of various process parameters in droplet-based printing relies on the proper design and stable operation of the inkjet nozzle. The setup of the experiment is illustrated in Fig.\ref{S7}.

For the experiment, we utilized the Droplet Shape Analyzer (DSA Inkjet) developed by Krüss (Germany) and the MH5420 inkjet nozzle manufactured by RICOH.

The goal was to control the driving voltage ${V}_\text{t}$ to produce a continuous stream of droplets. A high-speed camera was employed to precisely capture the morphology of the droplets at each moment. The experimental ink has a density of 1250 $\text{Kg/}{\text{m}^{3}}$, a viscosity of 10 cP and a surface tension coefficient of 0.04 $\text{N/m}$. 

In this study, the design parameter of interest was the driving voltage, ${V}_\text{t}$, which was varied across the range ${{V}_\text{t}}\in [15,30]\ \text{V}$. Settings for ID-DMD identification are provided in Tab.\ref{tab.S6}. The experimental results for a driving voltage of ${{V}_{\text{t}}}=21\ \text{V}$ are shown in Tab.I.

\linespread{1.2}
\begin{table*}[!ht] 
    \centering    
\noindent
\caption{ID-DMD settings for the droplet test}
\label{tab.S6}

    \begin{tabular}{|p{4cm}|p{9cm}|}

\hline
\makecell[l] {Training parameter} & \makecell[l] {${V}_\text{t}=\left\{18, 24, 30 \right\}$} \\

\hline
\makecell[l] {Time period} & \makecell[l] {$t\in [0,1]\times {10}^{-4}\ \text{s}$} \\

\hline
\makecell[l] {Sampling time} & \makecell[l] {$\Delta t=1\ \upmu \text{s}$} \\

\hline
\makecell[l] {Hyper-parameters} & \makecell[l] {${r}_\text{Z}={r}_{\Xi}=200$} \\

\hline
\makecell[l] {Scaling factor for ${V}_\text{t}$} & \makecell[l] {$\alpha=0.01$} \\

\hline
\makecell[l] {ID-DMD} & \makecell[l] {$\mathbf{x}_{k}=(\mathbf{A}_{0}+{{V}_\text{t}}{\mathbf{A}_{1}}){\mathbf{x}_{k-1}}$} \\

\hline
\end{tabular}
\end{table*}
\linespread{1}

\section{Section 4: Comparison with advanced machine learning methods}
For the Burgers' equation, the settings for the ID-DMD identification are provided in Tab.\ref{tab.S7}

\linespread{1.2}
\begin{table*}[!ht] 
    \centering    
\noindent
\caption{ID-DMD settings for the Burgers' equation}
\label{tab.S7}

    \begin{tabular}{|p{4cm}|p{9cm}|}

\hline
\makecell[l] {Training parameter} & \makecell[l] {$v=\left\{0.014,0.022,0.030,0.038,0.046 \right\}$} \\

\hline
\makecell[l] {Time period} & \makecell[l] {$t\in [0,1]\ \text{s}$}\\

\hline
\makecell[l] {Sampling time} & \makecell[l] {$\Delta t=0.01\ \text{s}$}\\

\hline
\makecell[l] {Hyper-parameters} & \makecell[l] {${r}_\text{Z}={r}_{\Xi}=40$} \\

\hline
\makecell[l] {Scaling factor for ${V}_\text{t}$} & \makecell[l] {$\alpha=1$} \\

\hline
\makecell[l] {ID-DMD} & \makecell[l] {$\mathbf{x}_{k}=(\mathbf{A}_{0}+{v}{\mathbf{A}_{1}}){\mathbf{x}_{k-1}}$} \\

\hline
\end{tabular}
\end{table*}
\linespread{1}

\begin{figure}[!ht]
  \centering
  \includegraphics[width=0.9\textwidth]{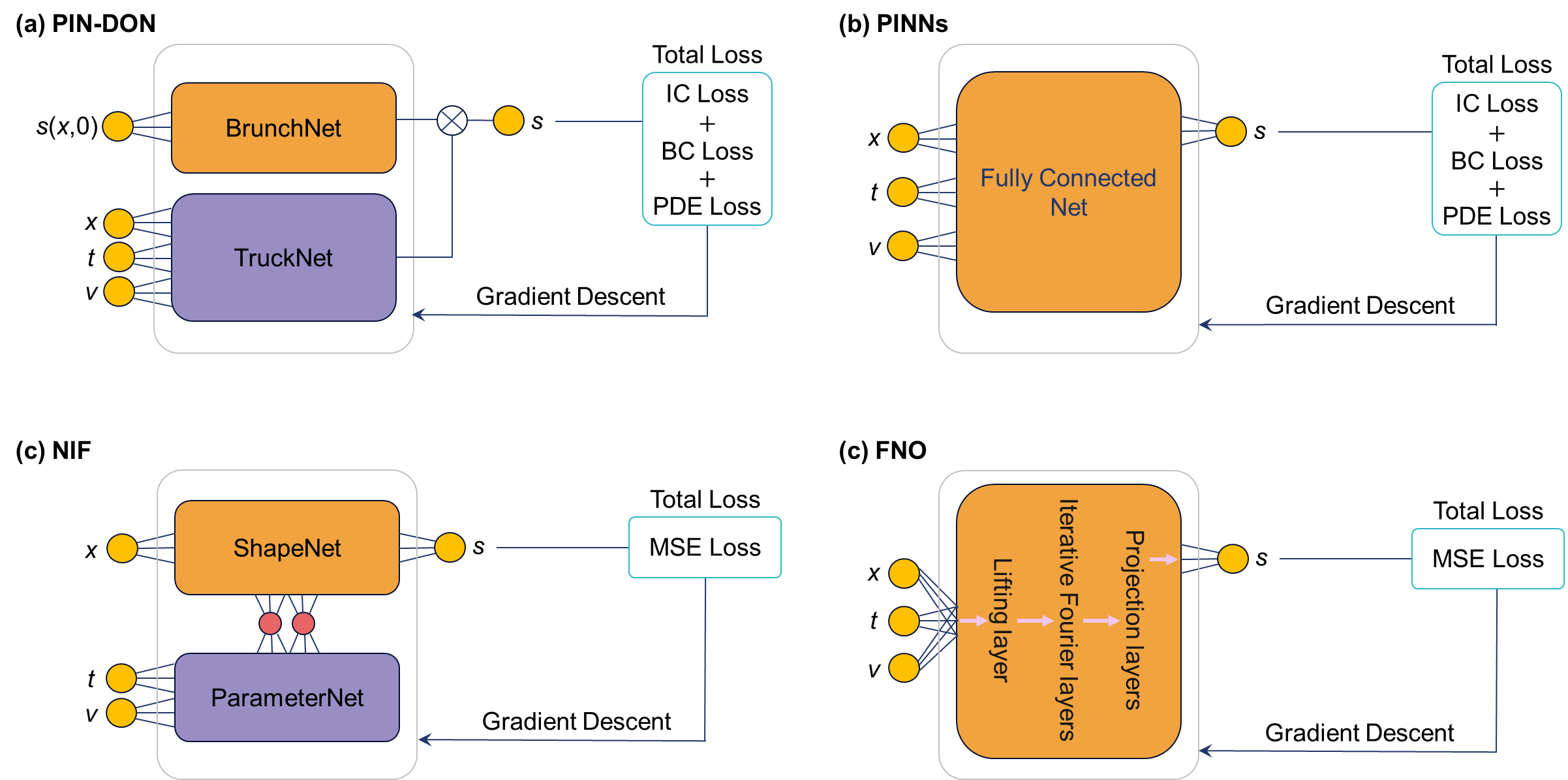}
  \caption{
  Extended architectures of parametric PI-DON, PINNs, NIF, and FNO.
  }
  \label{S8}
\end{figure}

To compare ID-DMD with advanced machine learning approaches, including Physics-Informed DeepONet (PI-DON), Physics-Informed Neural Networks (PINNs), Neural Implicit Flow (NIF), and Fourier Neural Operator (FNO), the same original training dataset $[x,t,v,s(x,t,v)]$ is used. The dataset is structured as a three-dimensional grid, where the spatial coordinates $x$ range from 0 to 1 in steps of 0.01, and the viscosity parameters $v$ take five specific values listed in Tab.\ref{tab.S7}, resulting in a dataset with dimensions $101\times 101\times 5$ and a total of 51015 data points. The Initial Conditions (IC), defined at $t=0$, consist of 505 data points ($101\times 5$), while the Boundary Conditions (BC), specified at $x=0$ and $x=1$, contribute 1010 data points ($101\times 5\times 2$). The extended architectures of parametric PI-DON, PINNs, NIF, and FNO, adapted to this dataset structure, are detailed in Fig.\ref{S8}, ensuring a consistent and fair comparison across methods.

\textbf{PI-DON} employs a dual-network architecture to process the input data, with distinct roles for the branch and trunk networks. The branch network is designed to capture the amplitudes $s(x, 0)$, corresponding to the initial conditions. We randomly sampled 2500 points from the remaining spatiotemporal domain for the training process. Simultaneously, the trunk network takes as input a matrix comprising the spatial coordinates $x$, time $t$, and viscosity parameter $v$, enabling the model to process the complete parametric information. The training process is governed by a physics-informed loss function, which incorporates residuals of the governing PDE along with terms ensuring adherence to the initial and boundary conditions. This ensures that the model respects the underlying physical laws during training. The PI-DON model contains 153000 trainable parameters and is trained over 20000 epochs using an initial learning rate of 0.001, which decays by a factor of 0.9 every 2000 epochs. This setup balances learning efficiency with model convergence, enabling the network to accurately capture the dynamics of the system.

\textbf{PINNs} take spatial coordinates $x$, time $t$, and viscosity $v$ as inputs, with the corresponding output data $s$ paired for supervised learning. The loss function integrates three key components: the IC, BC, and the residuals of the governing PDE. This ensures that the predicted solution remains consistent with the underlying physics and satisfies the specified constraints. For training, the input dataset consists of 100 randomly selected points from the IC and BC, combined with 10000 randomly sampled points from the interior spatiotemporal domain. These points are organized into a compact input matrix, enabling efficient processing. PINNs employ a minimalistic architecture with only 3041 trainable parameters, making the model lightweight and computationally efficient. Training is conducted over 20000 epochs with a fixed learning rate of 0.1, ensuring stable convergence while preserving computational.

\textbf{FNO} processes input data in a grid-based format, where each row of the input matrix consists of spatial coordinates $x$, time $t$, viscosity $v$, and the corresponding output $s$. This structured input allows FNO to operate effectively across both spatial and temporal domains, leveraging its unique ability to model complex operator mappings. The FNO employs Fourier transforms to represent and learn these mappings in the frequency domain. This approach not only makes the model resolution-invariant, allowing it to generalize across different grid sizes, but also enhances computational efficiency. The model’s loss function is designed to minimize the Mean Squared Error (MSE) between its predictions and the ground truth, ensuring accurate learning of the underlying dynamics.
In this example, the FNO features 52512001 trainable parameters. This large capacity allows it to capture the intricate dynamics of PDEs on discretized grids. The model is trained over 20000 epochs with a learning rate of 0.0005, enabling it to achieve stable convergence and precise results.

\textbf{NIF} utilizes an input format that separates spatial dependencies from parametric dependencies, enabling efficient modeling of complex systems. The architecture is composed of two key components: ShapeNet and ParameterNet. ShapeNet processes spatial coordinates $x$, effectively capturing spatial features, while ParameterNet encodes the temporal $t$ and viscosity $v$ parameters into a compact matrix format. This decoupling allows the model to efficiently handle spatiotemporal variations and parametric influences. The loss function is designed to minimize the MSE between the model’s predictions and the ground truth output $s$. Unlike physics-informed approaches, NIF focuses solely on data-driven accuracy, making it simple and computationally lightweight. With only 5883 trainable parameters, NIF is highly efficient and requires significantly less computational resources compared to larger models. The NIF is trained for 40000 epochs using a fixed learning rate of 0.001. This extended training period ensures convergence and enhances the model's ability to perform scalable and efficient dimensionality reduction for large-scale spatiotemporal problems.

The training details for each model are summarized in Tab.\ref{tab.S8}.

\linespread{1.2}
\begin{table*}[!ht] 
    \centering    
\noindent
\caption{Training details for advanced machine learning approaches}
\label{tab.S8}

    \begin{tabular}{|p{2.5cm}|p{4cm}|p{5cm}|p{2cm}|p{2.5cm}|}

\hline
\makecell[c] {Methods} & \makecell[c] {Number of trainable \\ parameters} & \makecell[c]{Learning rate} & \makecell[c]{Total epoch}& \makecell[c]{Loss function}\\

\hline
\makecell[c] {PI-DON} & \makecell[c] {153,000} & \makecell[c] {Initial learning rate of 0.001, \\which decays by a factor of 0.9 \\every 2,000 steps} & \makecell[c] {20,000} & \makecell[c] {BC,IC,\\ PDE residual}\\

\hline
\makecell[c] {PINNs} & \makecell[c] {3,041} & \makecell[c] {0.1} & \makecell[c] {20000} & \makecell[c] {BC,IC,\\ PDE residual}\\

\hline
\makecell[c] {FNO} & \makecell[c] {52,512,001} & \makecell[c] {0.0005} & \makecell[c] {20,000} & \makecell[c] {MSE}\\

\hline
\makecell[c] {NIF} & \makecell[c] {5,883} & \makecell[c] {0.001} & \makecell[c] {20,000} & \makecell[c] {MSE}\\

\hline
\end{tabular}
\end{table*}
\linespread{1}

\section{Section 5: Physical interpretability}
\subsection{Linear building structure}
\begin{figure}[!ht]
  \centering
  \includegraphics[width=0.5\textwidth]{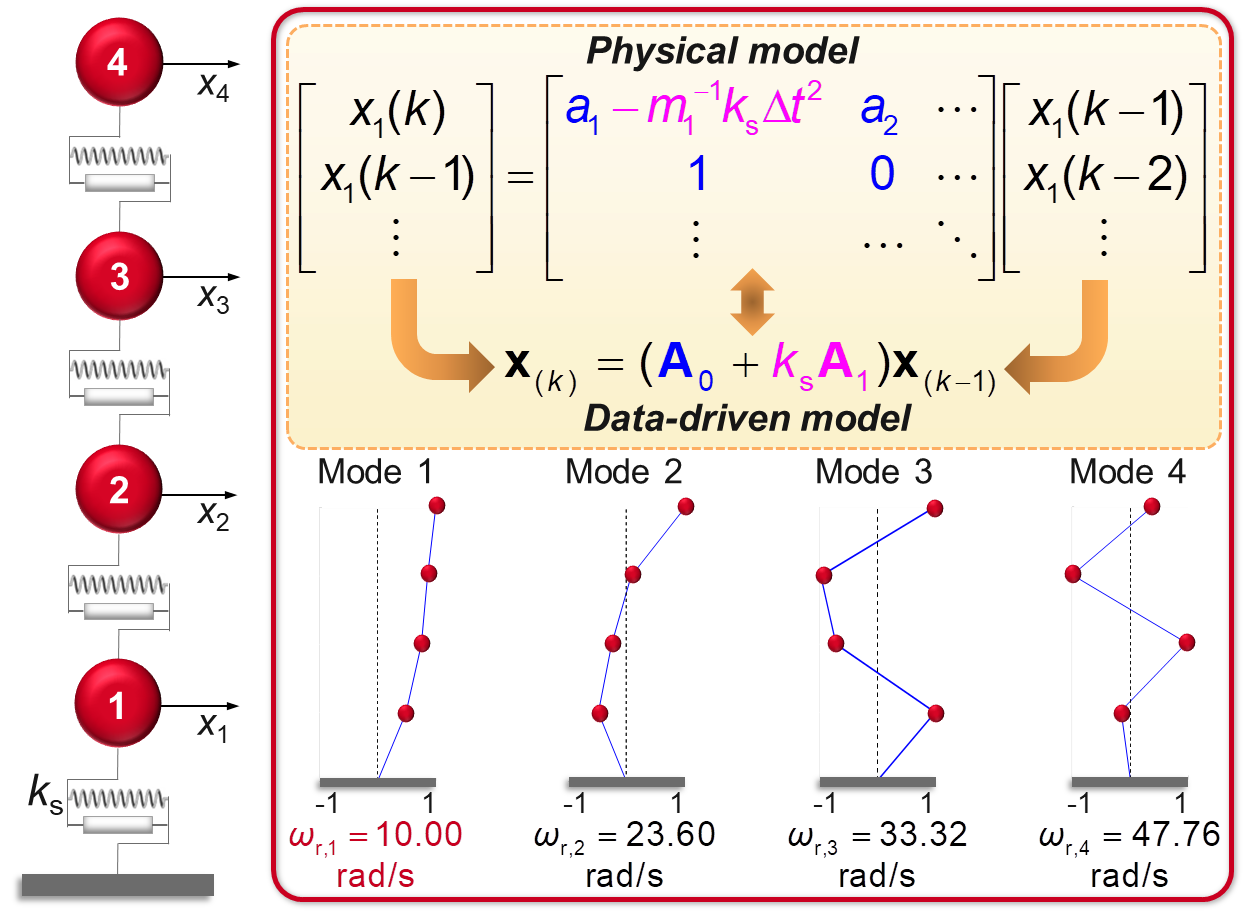}
  \caption{
  The 4-DoF linear building system and pole placement design using the ID-DMD.
  }
  \label{S9}
\end{figure}

This section demonstrates how the ID-DMD can effectively represent complex dynamic systems while maintaining physical interpretability. The ID-DMD is inherently transparent and physically interpretable. Fig.\ref{S9} highlights the natural connections between the data-driven ID-DMD and a physical state-space model, using a 4-Degree-of-Freedom (4-DoF) linear building system as an example. The characteristic parameters of the building are detailed in Section III.A. Here, the bottom linear stiffness ${k}_\text{s}={k}_{1}$ is selected as the design parameter.

The physical state space model of the linear building system can be achieved from the ODE with $\mathbf{F}_\text{non}=\mathbf{0}$ in equation\ref{eqS21}:

\begin{equation}
    \left\{\begin{aligned}
    & {{m}_{1}}{\ddot{x}_{1}}+({c}_{1}+{c}_{2}){\dot{x}_{1}}-{{c}_{2}}{\dot{x}_{2}}+({k}_{1}+{k}_{2}){{x}_{1}}-{{k}_{2}}{{x}_{2}}=0 \\ 
    & {{m}_{2}}{\ddot{x}_{2}}-{{c}_{2}}{\dot{x}_{1}}+({c}_{2}+{c}_{3}){\dot{x}_{2}}-{{c}_{3}}{\dot{x}_{3}}-{{k}_{2}}{{x}_{1}}+({k}_{2}+{k}_{3}){{x}_{2}}-{{k}_{3}}{{x}_{3}}=0 \\ 
    & {{m}_{3}}{\ddot{x}_{3}}-{{c}_{3}}{\dot{x}_{2}}+({c}_{3}+{c}_{4}){\dot{x}_{3}}-{{c}_{4}}{\dot{x}_{4}}-{{k}_{3}}{{x}_{2}}+({k}_{3}+{k}_{4}){{x}_{3}}-{{k}_{4}}{{x}_{4}}=0 \\ 
    & {{m}_{4}}{\ddot{x}_{4}}-{{c}_{4}}{\dot{x}_{3}}+{{c}_{4}}{\dot{x}_{4}}-{{k}_{4}}{{x}_{3}}+{{k}_{4}}{{x}_{4}}=0 \\ 
    \end{aligned} \right. \label{eqS25}
\end{equation}

By using the central difference method

\begin{equation*}
    \dot{x}=\frac{x(k)-x(k-1)}{\Delta t}\ \text{and}\ \ddot{x}=\frac{x(k+1)-2x(k)+x(k-1)}{\Delta {t}^{2}}
\end{equation*}

\noindent there is

\begin{equation}
    \left\{\begin{aligned}
    & {{m}_{1}}{{x}_{1}}(k)=[2{{m}_{1}}-({c}_{1}+{c}_{2})\Delta t-({k}_{1}+{k}_{2})\Delta {t}^{2}]{{x}_{1}}(k-1)+[-{m}_{1}+({c}_{1}+{c}_{2})\Delta t]{{x}_{1}}(k-2) \\ 
    & \ \ \ \ +[{c}_{2}\Delta t+{k}_{2}\Delta {t}^{2}]{{x}_{2}}(k-1)+\Delta t{{c}_{2}}{{x}_{2}}(k-2) \\ 
    & {{m}_{2}}{{x}_{2}}(k)=[{c}_{2}\Delta t+{k}_{2}\Delta {t}^{2}]{{x}_{1}}(k-1)-{c}_{2}\Delta t{{x}_{1}}(k-2) \\ 
    & \ \ \ \ +[2{{m}_{2}}-({c}_{2}+{c}_{3})\Delta t-({k}_{2}+{k}_{3})\Delta {t}^{2}]{{x}_{2}}(k-1)+[-{m}_{2}+({c}_{2}+{c}_{3})\Delta t]{{x}_{2}}(k-2) \\ 
    & \ \ \ \ +[{c}_{3}\Delta t+{k}_{3}\Delta {t}^{2}]{{x}_{3}}(k-1)-{c}_{3}\Delta t{{x}_{3}}(k-2) \\ 
    & {{m}_{3}}{{x}_{3}}(k)=[{c}_{3}\Delta t+{k}_{3}\Delta {t}^{2}]{{x}_{2}}(k-1)-{c}_{3}\Delta t{{x}_{2}}(k-2) \\ 
    & \ \ \ \ +[2{m}_{3}-({c}_{3}+{c}_{4})\Delta t-({k}_{3}+{k}_{4})\Delta {t}^{2}]{{x}_{3}}(k-1)+[-{m}_{3}+({c}_{3}+{c}_{4})\Delta t]{{x}_{3}}(k-2) \\ 
    & \ \ \ \ +[{c}_{4}\Delta t+{k}_{4}\Delta {t}^{2}]{{x}_{4}}(k-1)-{c}_{4}\Delta t{{x}_{4}}(k-2) \\ 
    & {{m}_{4}}{{x}_{4}}(k)=[2{m}_{4}-{c}_{4}\Delta t-{k}_{4}\Delta {t}^{2}]{{x}_{4}}(k-1)+[-{m}_{4}+{c}_{4}\Delta t]{{x}_{4}}(k-2) \\ 
    & \ \ \ \ +[{c}_{4}\Delta t+{k}_{4}\Delta {t}^{2}]{{x}_{3}}(k-1)-{c}_{4}\Delta t{{x}_{3}}(k-2) \\ 
    \end{aligned} \right. \label{eqS26}
\end{equation}

\noindent which can be formulated as a matrix representation

\begin{equation}
    \left[\begin{matrix}
    {{x}_{1}}(k) \\
    {{x}_{1}}(k-1) \\
    {{x}_{2}}(k) \\
    \vdots \\
    \end{matrix} \right]=\left[ \begin{matrix}
    {a}_{1}-{{m}_{1}}^{-1}{k}_{s}\Delta {t}^{2} & {a}_{2} & {a}_{3} & \cdots \\
    1 & 0 & 0 & \cdots \\
    0 & 0 & {a}_{4} & \cdots \\
    \vdots  & \vdots  & \vdots  & \ddots \\
    \end{matrix} \right]\left[\begin{matrix}
    {{x}_{1}}(k-1) \\
    {{x}_{1}}(k-2) \\
    {{x}_{2}}(k-1) \\
    \vdots \\
    \end{matrix} \right] \label{eqS27}
\end{equation}

\noindent where ${a}_{1},{a}_{2},{a}_{3},{a}_{4},...$ are constants.

From the physical model to the data-driven model, the system’s structure remains consistent, highlighting the transparency and physical interpretability of the ID-DMD approach. The state vectors $\mathbf{x}_{k}$ are formulated from the building’s responses ${{x}_{1}}(k),{{x}_{2}}(k),{{x}_{3}}(k),{{x}_{4}}(k)$, along with their time-delay embeddings ${{x}_{i}}(k-1),{{x}_{i}}(k-2),...$, where $i=1,2,3,4$. Settings for ID-DMD identification are listed in Tab.\ref{tab.S9}. 

\linespread{1.2}
\begin{table*}[!ht] 
    \centering    
\noindent
\caption{ID-DMD settings for the linear building}
\label{tab.S9}

    \begin{tabular}{|p{4cm}|p{9cm}|}

\hline
\makecell[l] {Training parameter} & \makecell[l] {${k}_\text{s}=\left\{1, 2, 3, 3.5 \right\}\times 10^9$} \\

\hline
\makecell[l] {Time period} & \makecell[l] {$t\in [0,8000]\ \text{s}$} \\

\hline
\makecell[l] {Sampling time} & \makecell[l] {$\Delta t=1/80\ \text{s}$} \\

\hline
\makecell[l] {Hyper-parameters} & \makecell[l] {${r}_\text{Z}={r}_{\Xi}=16$} \\

\hline
\makecell[l] {Scaling factor for ${c}_{3}$} & \makecell[l] {$\alpha=1\times 10^{-9}$} \\

\hline
\makecell[l] {ID-DMD} & \makecell[l] {$\mathbf{x}_{k}=({\mathbf{A}_{0}}+{{k}_\text{s}}{\mathbf{A}_{1}})\mathbf{x}_{k-1}$} \\

\hline
\end{tabular}
\end{table*}
\linespread{1}

This physical interpretability is demonstrated through pole placement in a 4-DoF linear building system. The desired first resonant frequency is set to $\omega_{\text{r,}1}=10\ \text{rad/s}$.  The ID-DMD-based design yields a linear stiffness of  ${k}_\text{s}=2.82\times {10}^{9}\ \text{N}/\text{m}$ $\pm 0.14\times {10}^{7}$. Key physical properties, such as mode shapes and higher-order resonant frequencies, are accurately evaluated as shown in Fig.\ref{S9}.

\subsection{Nonlinearly damped ODE}
In this section, we explore the ID-DMD modeling of a nonlinearly damped differential equation

\begin{equation}
    \frac{{\text{d}^{2}}y}{\text{d}{t}^{2}}+0.03\frac{\text{d}y}{\text{d}t}+100y+{{c}_{3}}(\frac{\text{d}y}{\text{d}t})^{3}=0 \label{eqS28}
\end{equation}

\noindent with the initial condition $x(0)=0.01$ and $\dot{x}(0)=0$, where ${c}_{3}$ represents the nonlinear damping and serves as the design parameter with ${c}_{3}\in [1,20]$. The ID-DMD representation of the nonlinearly damped differential equation is $\psi(\mathbf{x}_{k})=(\mathbf{A}_{\kappa,0}+{{c}_{3}}{\mathbf{A}_{\kappa,1}})\psi(\mathbf{x}_{k-1})$, where the observables are obtained by the polynomial projection as

\begin{equation}
    \psi (\mathbf{x}_{k})={{[y(k),\ y(k-1),{{y}^{2}}(k),\ y(k)y(k-1),{{y}^{2}}(k-1),\ldots]}^{\text{T}}} \label{eqS29}
\end{equation}

\noindent up to the 8th order with a total number of 45 observables. Settings for ID-DMD identification are provided in Tab.\ref{tab.S10}.

\linespread{1.2}
\begin{table*}[!ht] 
    \centering    
\noindent
\caption{ID-DMD settings for the nonlinearly damped ODE}
\label{tab.S10}

    \begin{tabular}{|p{4cm}|p{9cm}|}

\hline
\makecell[l] {Training parameter} & \makecell[l] {${c}_{3}=\left\{1, 10, 20 \right\}$} \\

\hline
\makecell[l] {Time period} & \makecell[l] {$t\in [0,200]\ \text{s}$} \\

\hline
\makecell[l] {Sampling time} & \makecell[l] {$\Delta t=1/32\ \text{s}$} \\

\hline
\makecell[l] {Hyper-parameters} & \makecell[l] {${r}_\text{Z}={r}_{\Xi}=35$} \\

\hline
\makecell[l] {Scaling factor for ${c}_{3}$} & \makecell[l] {$\alpha=0.1$} \\

\hline
\makecell[l] {ID-DMD} & \makecell[l] {$\psi(\mathbf{x}_{k})=(\mathbf{A}_{\kappa,0}+{{c}_{3}}{\mathbf{A}_{\kappa,1}})\psi(\mathbf{x}_{k-1})$} \\

\hline
\end{tabular}
\end{table*}
\linespread{1}

Dominant modes and their corresponding characteristic frequencies are extracted using the ID-DMD method, as shown in Fig.3. In Fig.3(a), the characteristic frequencies are evaluated across different design parameters. Frequencies that remain stable correspond to the true dominant modes, while those that vary with design parameters are identified as spurious modes. Compared to the stable dominant modes, spurious modes typically exhibit larger damping coefficients (i.e., larger real parts of the eigenvalues), causing their influence on system responses to decay rapidly. 

In Fig.3 (a) and (b), it is observed that the first mode originates from the linear states ($y(k)$, $y(k-1)$), corresponding to the characteristic frequency $\omega_\text{e}=10\ \text{rad/s}$. This can be easily verified via the ODE\ref{eqS28}, where the linear natural frequency is calculated as ${\omega_\text{r}}=10\ \text{rad/s}$.  The second mode arises from the squared projections of the states (${{y}^{2}}(k)$, $y(k)y(k-1)$, ${{y}^{2}}(k-1)$), occurring at the second-order modulation frequency ${\omega_\text{e}}=20\ \text{rad/s}$, which is twice the first-order frequency. The third stable mode appears at ${\omega_\text{e}}=30\ \text{rad/s}$, resulting from contributions of the linear states and their cubic projections (${{y}^{3}}(k)$, ${{y}^{2}}(k)y(k-1)$, $y(k){{y}^{2}}(k-1)$, ${{y}^{3}}(k-1)$). As illustrated in Fig.3(b) (iii), the modes associated with cubic projection states decrease as the nonlinear damping ${c}_{3}$ increases.

Figs.3(b) and (c) illustrate the frequency modulation characteristics of nonlinear systems~\cite{lang1996output}. In Fig.3(b), the linear states, representing the system output responses, contribute to both the first-order and third-order modes. As a result, when the system is excited by a single-tone input $u(t)=\cos(10t)$, the output response in Fig.3(c) contains the fundamental frequency and a third-order modulation component.

By introducing nonlinear damping, the specific energy loss ${E}_\text{d}=\sum\nolimits_{k=1}^{K}{{{\left| y(k) \right|}^{2}}}$ can be controlled without altering the system's settling time. As shown in Fig.3(d), an energy loss of ${{E}_\text{d}}<0.013$ is achieved when ${c}_{3}>6$. These properties are widely utilized in engineering applications for vibration control, but cannot be achieved by using linear damping.

\section*{References}
\bibliographystyle{unsrt}
\bibliography{mainNotes,mainNotes_SI}

\end{document}